\documentclass[twoside,twocolumn,9pt]{article}
\usepackage{extsizes}
\usepackage[super,sort&compress,comma]{natbib} 
\usepackage[version=3]{mhchem}
\usepackage[left=1.5cm, right=1.5cm, top=1.785cm, bottom=2.0cm]{geometry}
\usepackage{balance}
\usepackage{mathptmx}
\usepackage{sectsty}
\usepackage{graphicx} 
\usepackage{lastpage}
\usepackage[format=plain,justification=justified,singlelinecheck=false,font={stretch=1.125,small,sf},labelfont=bf,labelsep=space]{caption}
\usepackage{float}
\usepackage{fancyhdr}
\usepackage{fnpos}
\usepackage[english]{babel}
\addto{\captionsenglish}{%
  
}
\usepackage{array}
\usepackage{droidsans}
\usepackage{charter}
\usepackage[T1]{fontenc}
\usepackage{setspace}
\usepackage[compact]{titlesec}
\usepackage{hyperref}
\usepackage[usenames,dvipsnames]{color} 
\usepackage{graphicx}%

\usepackage{amsmath,amssymb,amsfonts}%
\usepackage{amsthm}%
\usepackage{mathrsfs}%
\usepackage{textcomp}%
\usepackage{manyfoot}%
\usepackage{booktabs}%
\usepackage{algorithm}%
\usepackage{algorithmicx}%
\usepackage{algpseudocode}%
\usepackage{listings}%

\usepackage{pifont}% http://ctan.org/pkg/pifont

\usepackage[table,xcdraw]{xcolor}
\usepackage{multirow}%

\newcommand{\ourmodel}{nach0}

\definecolor{cream}{RGB}{222,217,201}

\begin{document}

\pagestyle{fancy}
\thispagestyle{plain}
\fancypagestyle{plain}{
%%%HEADER%%%
\renewcommand{\headrulewidth}{0pt}
}
%%%END OF HEADER%%%

%%%PAGE SETUP - Please do not change any commands within this section%%%
\makeFNbottom
\makeatletter
\renewcommand\LARGE{\@setfontsize\LARGE{15pt}{17}}
\renewcommand\Large{\@setfontsize\Large{12pt}{14}}
\renewcommand\large{\@setfontsize\large{10pt}{12}}
\renewcommand\footnotesize{\@setfontsize\footnotesize{7pt}{10}}
\makeatother

\renewcommand{\thefootnote}{\fnsymbol{footnote}}
\renewcommand\footnoterule{\vspace*{1pt}% 
\color{cream}\hrule width 3.5in height 0.4pt \color{black}\vspace*{5pt}} 
\setcounter{secnumdepth}{5}

\makeatletter 
\renewcommand\@biblabel[1]{#1}            
\renewcommand\@makefntext[1]% 
{\noindent\makebox[0pt][r]{\@thefnmark\,}#1}
\makeatother 
\renewcommand{\figurename}{\small{Fig.}~}
\sectionfont{\sffamily\Large}
\subsectionfont{\normalsize}
\subsubsectionfont{\bf}
\setstretch{1.125} %In particular, please do not alter this line.
\setlength{\skip\footins}{0.8cm}
\setlength{\footnotesep}{0.25cm}
\setlength{\jot}{10pt}
\titlespacing*{\section}{0pt}{4pt}{4pt}
\titlespacing*{\subsection}{0pt}{15pt}{1pt}
%%%END OF PAGE SETUP%%%

%%%FOOTER%%%
\fancyfoot{}
\fancyfoot[LO,RE]{\vspace{-7.1pt}\includegraphics[height=9pt]{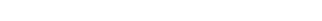}}
\fancyfoot[CO]{\vspace{-7.1pt}\hspace{13.2cm}\includegraphics{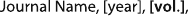}}
\fancyfoot[CE]{\vspace{-7.2pt}\hspace{-14.2cm}\includegraphics{head_foot/RF}}
\fancyfoot[RO]{\footnotesize{\sffamily{1--\pageref{LastPage} ~\textbar  \hspace{2pt}\thepage}}}
\fancyfoot[LE]{\footnotesize{\sffamily{\thepage~\textbar\hspace{3.45cm} 1--\pageref{LastPage}}}}
\fancyhead{}
\renewcommand{\headrulewidth}{0pt} 
\renewcommand{\footrulewidth}{0pt}
\setlength{\arrayrulewidth}{1pt}
\setlength{\columnsep}{6.5mm}
\setlength\bibsep{1pt}
%%%END OF FOOTER%%%

%%%FIGURE SETUP - please do not change any commands within this section%%%
\makeatletter 
\newlength{\figrulesep} 
\setlength{\figrulesep}{0.5\textfloatsep} 

\newcommand{\topfigrule}{\vspace*{-1pt}% 
\noindent{\color{cream}\rule[-\figrulesep]{\columnwidth}{1.5pt}} }

\newcommand{\botfigrule}{\vspace*{-2pt}% 
\noindent{\color{cream}\rule[\figrulesep]{\columnwidth}{1.5pt}} }

\newcommand{\dblfigrule}{\vspace*{-1pt}% 
\noindent{\color{cream}\rule[-\figrulesep]{\textwidth}{1.5pt}} }

\makeatother
%%%END OF FIGURE SETUP%%%

%%%TITLE, AUTHORS AND ABSTRACT%%%
\twocolumn[
  \begin{@twocolumnfalse}
{\includegraphics[height=30pt]{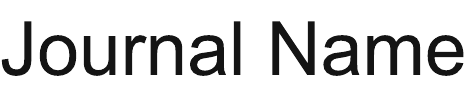}\hfill\raisebox{0pt}[0pt][0pt]{\includegraphics[height=55pt]{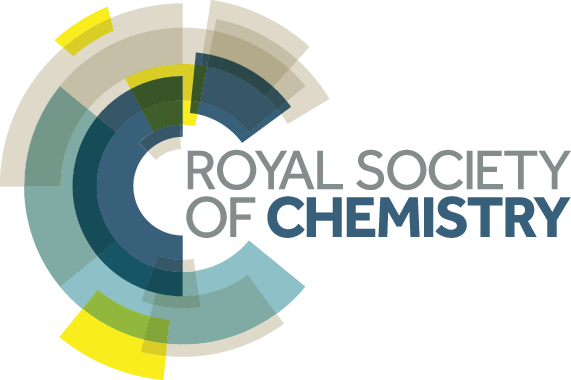}}\\[1ex]
\includegraphics[width=18.5cm]{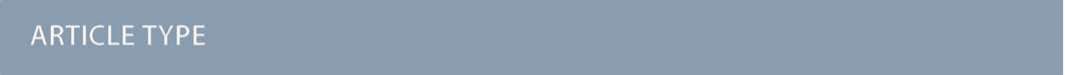}}\par
\vspace{1em}
\sffamily
\begin{tabular}{m{4.5cm} p{13.5cm} }

\includegraphics{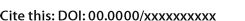} & \noindent\LARGE{\textbf{\ourmodel{}: Multimodal Natural and Chemical Languages Foundation Model$^\dag$}} \\%Article title goes here instead of the text "This is the title"
\vspace{0.3cm} & \vspace{0.3cm} \\

 & \noindent\large{Micha Livne,\textit{$^{a\dag}$} Zulfat Miftahutdinov,\textit{$^{b\dag}$} Elena Tutubalina,\textit{$^{c\dag}$} Maksim Kuznetsov,\textit{$^{b\dag}$} Daniil Polykovskiy,\textit{$^{b}$} Annika Brundyn,\textit{$^{a}$} Aastha Jhunjhunwala,\textit{$^{a}$} Anthony Costa,\textit{$^{a}$} Alex Aliper,\textit{$^{d}$} Alán Aspuru-Guzik,\textit{$^{e}$} and Alex Zhavoronkov\textit{$^{c\ddag}$}} \\%Author names go here instead of "Full name", etc.

\includegraphics{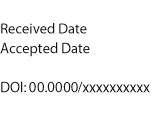} & \noindent\normalsize{Large Language Models (LLMs) have substantially driven scientific progress in various domains, and many papers have demonstrated their ability to tackle complex problems with creative solutions.
Our paper introduces a new foundation model, \ourmodel, capable of solving various chemical and biological tasks: biomedical question answering, named entity recognition, molecular generation, molecular synthesis, attributes prediction, and others.
\ourmodel{} is a multi-domain and multi-task encoder-decoder LLM pre-trained on unlabeled text from scientific literature, patents, and molecule strings to incorporate a range of chemical and linguistic knowledge. We employed instruction tuning, where specific task-related instructions are utilized to fine-tune \ourmodel{} for the final set of tasks.
To train \ourmodel{} effectively, we leverage the NeMo framework, enabling efficient parallel optimization of both base and large model versions. 
Extensive experiments demonstrate that our model outperforms state-of-the-art baselines on single-domain and cross-domain tasks. Furthermore, it can generate high-quality outputs in molecular and textual formats, showcasing its effectiveness in multi-domain setups.} \\%The abstrast goes here instead of the text "The abstract should be..."

\end{tabular}

 \end{@twocolumnfalse} \vspace{0.6cm}

  ]
%%%END OF TITLE, AUTHORS AND ABSTRACT%%%

%%%FONT SETUP - please do not change any commands within this section
\renewcommand*\rmdefault{bch}\normalfont\upshape
\rmfamily
\section*{}
\vspace{-1cm}

%%%FOOTNOTES%%%

\footnotetext{\textit{$^{a}$~NVIDIA, 2788 San Tomas Expressway, Santa Clara, 95051, CA, US}}
\footnotetext{\textit{$^{b}$~Insilico Medicine Canada Inc., 3710-1250 René-Lévesque west, Montreal, Quebec, Canada}}
\footnotetext{\textit{$^{c}$~Insilico Medicine Hong Kong Ltd., Unit 310, 3/F, Building 8W, Phase 2, Hong Kong Science Park, Pak Shek Kok, New Territories, Hong Kong}}
\footnotetext{\textit{$^{d}$~Insilico Medicine AI Ltd., Level 6, Unit 08, Block A, IRENA HQ Building, Masdar City, Abu Dhabi, United Arab Emirates}}
\footnotetext{\textit{$^{e}$~University of Toronto, Lash Miller Building 80 St. George Street, Toronto, Ontario, Canada. Email: alan@aspuru.com}}

%Please use \dag to cite the ESI in the main text of the article.
%If you article does not have ESI please remove the the \dag symbol from the title and the footnotetext below.
\footnotetext{\dag~These authors contributed equally to this work.}
\footnotetext{\ddag~Email: alex@insilicomedicine.com}
%additional addresses can be cited as above using the lower-case letters, c, d, e... If all authors are from the same address, no letter is required
%\footnotetext{\ddag~Additional footnotes to the title and authors can be included \textit{e.g.}\ `Present address:' or `These authors contributed equally to this work' as above using the symbols: \ddag, \textsection, and \P. Please place the appropriate symbol next to the author's name and include a \texttt{\textbackslash footnotetext} entry in the the correct place in the list.}

%%%END OF FOOTNOTES%%%

%%%MAIN TEXT%%%%
\section{Introduction}\label{sec1}

Large-scale pre-training of language models (LMs), such as BERT \citep{devlin-etal-2019-bert}, T5 \citep{raffel2020exploring}, BART \citep{lewis2020bart} and GPT \citep{brown2020language}, on vast amounts of text data has yielded impressive results on a variety of natural language processing (NLP) tasks. These models' success can be attributed to their ability to learn deeply contextualized representations of input tokens through self-supervision at scale \citep{devlin-etal-2019-bert}. Recently, foundation models have built upon the concept of self-supervised learning by pre-training a single model over unlabeled data that can be easily adapted to any task \citep{bommasani2021opportunities}. 

The application of neural network architectures and LMs has significantly advanced the field of chemistry, particularly in domain-specific information retrieval, drug development, and clinical trial design \citep{IEsearch,Miftahutdinov20213856,Miftahutdinov2021451,Tutubalina20206710,ct2023,putin2018adversarial,polykovskiy2018entangled,10.3389/fphar.2020.00269,aliper2016deep, kuznetsov2021molgrow}. These developments include neural molecular fingerprinting, generative approaches to small molecule design~\citep{putin2018adversarial,polykovskiy2018entangled,10.3389/fphar.2020.00269}, prediction of pharmacological properties, and drug repurposing~\citep{10.3389/fphar.2020.00269,aliper2016deep}. The clinical development of a drug is a time and money consuming process that typically requires several years and a billion-dollar budget to progress from phase 1 clinical trials to the patients~\citep{dowden2019trends}. The use of state-of-the-art neural network approaches and language models has the potential to facilitate the drug development process considerably.

\begin{figure}
    \centering
    \includegraphics[width=0.5\linewidth]{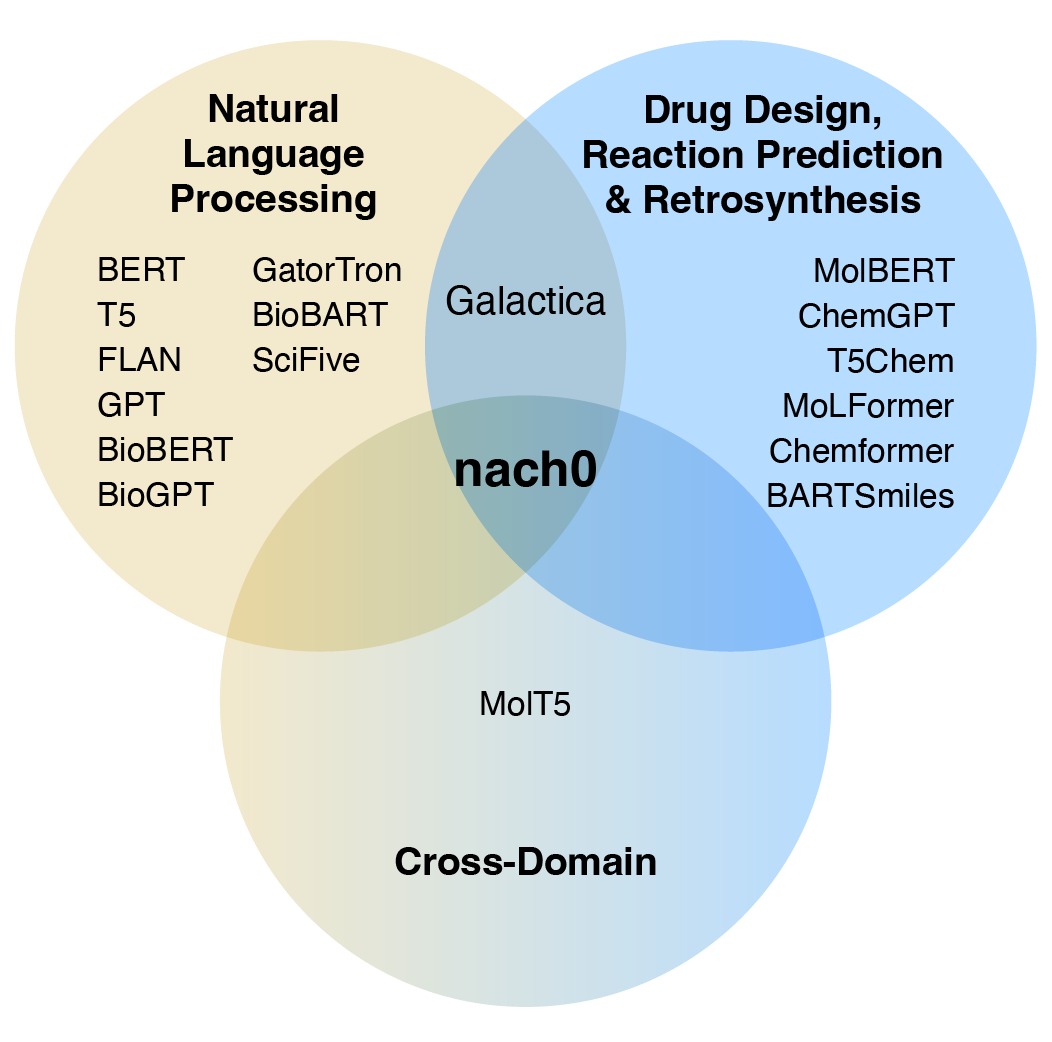}
    \caption{A Venn diagram that shows the relationships between fine-tuning data used in our study and related work. It is important to highlight that the majority of models typically treat the chemical space and the semantic space in the natural language domain independently. Novel cross-domain datasets such as Mol-Instructions \citep{fang2023mol} and MolT5 data \citep{edwards2022translation} have asked whether it is possible to unify representations of natural language and molecules for NLP and molecule generation tasks within a single model. In this work, we seek to answer this question. }
    \label{fig:venn}
\end{figure}

A number of LMs have been proposed for the biomedical domain, utilizing a variety of model families: for instance, researchers have developed BioBERT \citep{lee2020biobert}, based on BERT with 110 million parameters, and SciFive, based on T5-base and T5-large with 220 and 770 million parameters respectively, using biomedical literature from PubMed. NVIDIA has also developed BioMegatron models in the biomedical domain using a more extensive set of PubMed-derived free text, ranging from 345 million to 1.2 billion parameters. However, the datasets used in these models cover mainly biomedical natural language texts and contain biomedical named entities like drugs, genes, and cell lines names but omit important chemical structure descriptions in SMILES format. Enriching biomedical datasets with chemical structures is an important and challenging task. Recently, LMs such as Galactica \citep{taylor2022galactica}, based on Transformer architecture in a decoder-only setup \citep{vaswani2017attention} with 120 billion parameters in its largest setup, and MolT5 \citep{edwards2022translation}, based on T5-base and T5-large, were proposed to address this limitation. Both modes were pre-trained with natural language and chemical data, creating a shared representation space, yet were not fine-tuned on a diverse set of chemical tasks with instruction tuning in a multi-task fashion. The Venn diagram in Fig. \ref{fig:venn} provides a summary of the existing LMs. Furthermore, simple language models trained with molecular structures can reproduce complex molecular distributions \citep{flam2022language}, and even their 3D structure of molecules, materials and proteins using a GPT framework \citep{flam2023language}.

\begin{figure}
    \centering
    \includegraphics[width=1\linewidth]{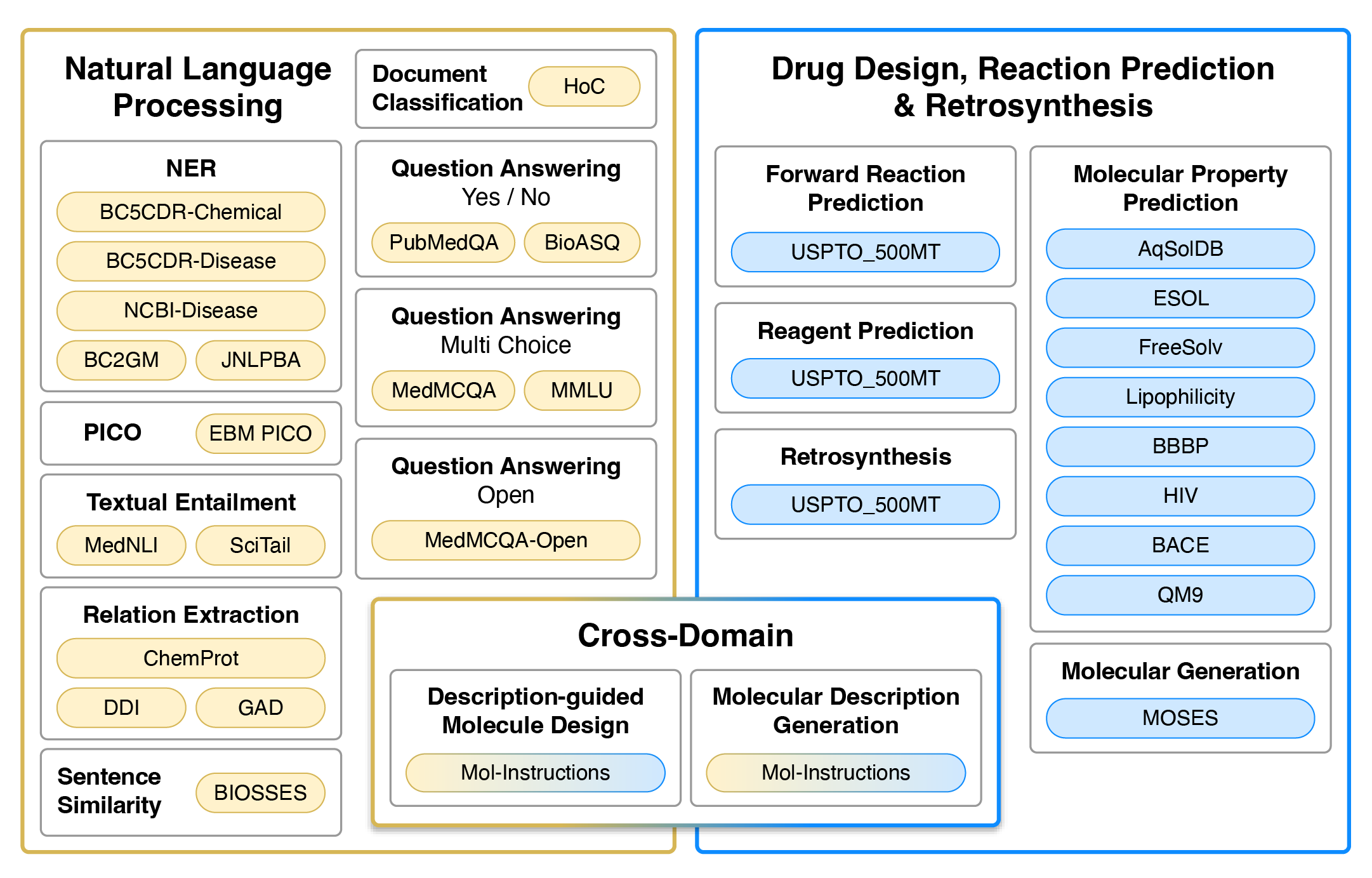}
    \caption{Datasets used for training and evaluation. Colour represents the type of tasks. Yellow and blue datasets are single-domain, typically requiring regression/classification losses or generation in the target domain (natural language or SMILES strings). Gradients from yellow to blue represent cross-domain generation tasks that require natural language input and SMILES output, or vise versa. }
    \label{fig:datasets}
\end{figure}

In this paper, we propose a unified encoder-decoder transformer named \ourmodel{} for natural language, chemical generalization and cross-domain tasks. We pre-train on both natural language and chemical data using Self Supervised Learning and employ \ourmodel{} as the foundation model for a wide range of downstream tasks (Fig. \ref{fig:datasets}). The tasks include well-known NLP problems such as information extraction, question answering, textual entailment, molecular structures and description generation, chemical property prediction, and reaction predictions. Inspired by \citet{raffel2020exploring,chung2022scaling}, we follow the intuition that tasks can be described via natural language instructions, such as “What reactants could be used to synthesize {\tt O=C(NC1CCN(Cc2ccccc2)CC1)c1c(Cl)cccc1[N+](=O)[O-]}” or “describe a molecule {\tt C1=CC(=CC=C1C[C@H](C(=O)[O-])N)O}”. Prompt design and instruction tuning are employed for model training using NVIDIA's Neural Modules (NeMo) framework \citep{kuchaiev2019nemo}, which provides scientists with a way to train and deploy LLMs using NVIDIA GPUs. Extensive evaluation in both in-domain and cross-domain setup demonstrates that \ourmodel{} is a powerful tool for the chemistry domain.  


\paragraph*{Contribution} 
Our contributions are three-fold:
\begin{enumerate}
    \item We introduce a biochemical foundation model \ourmodel{} and pre-train base and large versions of \ourmodel{} on molecular structures and textual data from scientific articles and patents. 
    \item We fine-tune \ourmodel{} in a supervised and multi-task manner, using a combination of diverse tasks specified through natural language prompts.
    \item Through the experimental validation on benchmark datasets, focusing on both single-domain and cross-domain tasks, we show that our model achieves competitive results with state-of-the-art encoder-decoder models specialized for single domain. 
\end{enumerate}
\section{Methods}\label{sec:methods}

\subsection{Framework \ourmodel{}}
\begin{figure}
    \centering
    \includegraphics[width=0.8\linewidth]{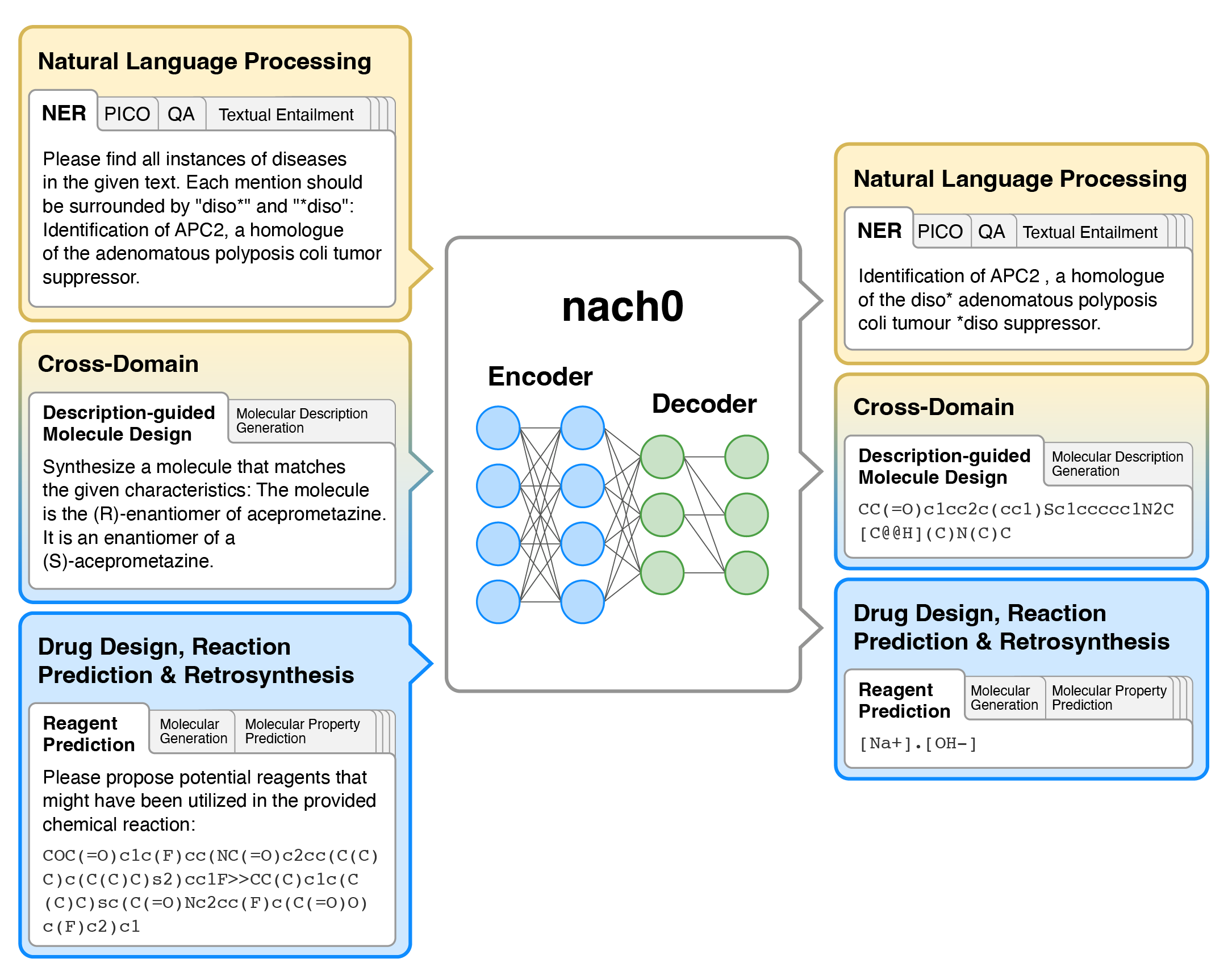}
    \caption{A diagram of \ourmodel{} which is a text-to-text framework. The model takes text as input and is trained to generate the desired target text for each specific task. This unified approach enables us to utilize the same model architecture, loss function, hyperparameters, and other components across our diverse range of mono-domain (NLP, CHEM) and cross-domain (NLP$\leftrightarrow$CHEM) tasks.}
    \label{fig:model}
\end{figure}

The aim of \ourmodel{} is to create a unified transformer capable of performing natural language, chemical generalization, and translation tasks simultaneously. Fig. \ref{fig:model} shows a diagram of our framework with several input/output examples.
The model's representations are learned from extensive and diverse chemical SMILES data and related textual data from scientific articles and patents. Similar to \citet{raffel2020exploring,chung2022scaling}, \ourmodel{} follows an encoder-decoder architecture that takes textual input and generates target responses. To train the model on a mixture of datasets partitioned into different tasks, we formulate all the tasks in a ``text-to-text'' format, where the model is given some text as a context or condition and produces the output in a text format. Each dataset is associated with multiple prompt templates used to format datasets' instances into input and target pairs. In particular, we train \ourmodel{} on three types of tasks (Fig. \ref{fig:datasets}):
\begin{itemize}
\item NLP tasks: named entity recognition (NER), PICO extraction, textual entailment, relation extraction, sentence similarity, document classification, question answering (yes/no, multi-choice, open);
\item chemistry-related (CHEM) tasks: molecular property prediction, molecular generation, forward reaction prediction, reagent prediction, retrosynthesis;
\item cross-domain (NLP$\leftrightarrow$CHEM) tasks: description-guided molecule design, molecular description generation;
\end{itemize}

% % Please add the following required packages to your document preamble:

% % Please add the following required packages to your document preamble:
% % \usepackage{booktabs}
% % \usepackage{multirow}
% % Please add the following required packages to your document preamble:
% % \usepackage{booktabs}
% % \usepackage{multirow}

\begin{table}[t!]
\small
\caption{List of datasets used in our study. We note that ESOL, FreeSolv, Lipophilicity, BBBP, HIV, BACE are included in the MoleculeNet benchmark~\citep{wu2018moleculenet}; QM9, MoleculeNet and USPTO\_500MT data are collected from Mol-Instructions \citep{fang2023mol}.}\label{tab:datalinks}
\begin{tabular}{|p{2.6cm}|c|c|p{1.4cm}|}
\hline
\textbf{Task} &
  \textbf{Dataset} &
  \textbf{Link} &
  \textbf{Train/Test split} \\\hline
\multirow{5}{*}{NER} &
  \multicolumn{1}{c|}{BC5CDR-Chemical \citep{li2016biocreative}} &
  \multicolumn{1}{c|}{\href{https://huggingface.co/datasets/bigbio/blurb/viewer/bc5chem}{link}} &
  predefined \\
 &
  \multicolumn{1}{c|}{BC5CDR-Disease \citep{li2016biocreative}} &
  \multicolumn{1}{c|}{\href{https://huggingface.co/datasets/bigbio/blurb/viewer/bc5disease}{link}} &
  predefined \\
 &
  \multicolumn{1}{c|}{NCBI-disease \citep{dougan2014ncbi}} &
  \multicolumn{1}{c|}{\href{https://huggingface.co/datasets/bigbio/blurb/viewer/ncbi_disease/}{link}} &
  predefined \\
 &
  \multicolumn{1}{c|}{BC2GM \citep{smith2008biocreative}} &
  \multicolumn{1}{c|}{\href{https://huggingface.co/datasets/bigbio/blurb/viewer/bc2gm}{link}} &
  predefined \\
 &
  \multicolumn{1}{c|}{JNLPBA \citep{collier2004introduction}} &
  \multicolumn{1}{c|}{\href{https://huggingface.co/datasets/bigbio/blurb/viewer/jnlpba}{link}} &
  predefined \\ \hline
\multicolumn{1}{|c|}{PICO} &
  \multicolumn{1}{c|}{EBM PICO \citep{nye2018corpus}} &
  \multicolumn{1}{c|}{\href{https://github.com/bigscience-workshop/biomedical}{link}} &
  predefined \\ \hline
\multirow{2}{*}{Textual Entailment} &
  \multicolumn{1}{c|}{MedNLI \citep{shivade2019mednli}} &
  \multicolumn{1}{c|}{\href{https://github.com/bigscience-workshop/biomedical}{link}} &
  predefined \\
 &
  \multicolumn{1}{c|}{SciTail \citep{khot2018scitail}} &
  \multicolumn{1}{c|}{\href{https://github.com/bigscience-workshop/biomedical}{link}} &
  predefined \\ \hline
\multirow{3}{*}{Relation Extraction} &
  \multicolumn{1}{c|}{ChemProt \citep{krallinger2017overview}} &
  \multicolumn{1}{c|}{\href{https://github.com/bigscience-workshop/biomedical}{link}} &
  predefined \\
&
  \multicolumn{1}{c|}{DDI \citep{herrero2013ddi}} &
  \multicolumn{1}{c|}{\href{https://github.com/bigscience-workshop/biomedical}{link}} &
  predefined \\
 &
  \multicolumn{1}{c|}{GAD \citep{bravo2015extraction}} &
  \multicolumn{1}{c|}{\href{https://github.com/bigscience-workshop/biomedical}{link}} &
  predefined \\ \hline
Sentence similarity &
  \multicolumn{1}{c|}{BIOSSES \citep{souganciouglu2017biosses}} &
  \multicolumn{1}{c|}{\href{https://github.com/bigscience-workshop/biomedical}{link}} &
  predefined \\ \hline
Document Classification &
  \multicolumn{1}{c|}{HoC \citep{hanahan2000hallmarks}} &
  \multicolumn{1}{c|}{\href{https://github.com/bigscience-workshop/biomedical}{link}} &
  predefined \\ \hline
\multirow{2}{\linewidth}{Question answering (Yes/No)} &
  \multicolumn{1}{c|}{PubMedQA \citep{jin2019pubmedqa}} &
  \multicolumn{1}{c|}{\href{https://github.com/bigscience-workshop/biomedical}{link}} &
  predefined \\ 
 &
  \multicolumn{1}{c|}{BioASQ \citep{nentidis2020results}} &
  \multicolumn{1}{c|}{\href{https://github.com/bigscience-workshop/biomedical}{link}} &
  predefined \\ \hline
\multirow{7}{\linewidth}{Molecular property prediction} &
  \multicolumn{1}{c|}{ESOL \citep{wu2018moleculenet}} &
  \multicolumn{1}{c|}{\multirow{6}{*}{\href{https://moleculenet.org}{link}}} &
  \multicolumn{1}{c|}{\multirow{6}{*}{predefined}} \\
 &
  \multicolumn{1}{c|}{FreeSolv \citep{wu2018moleculenet}} & &  \\
 &
  \multicolumn{1}{c|}{Lipophilicity \citep{wu2018moleculenet}} &  & \\
 &
  \multicolumn{1}{c|}{BBBP \citep{wu2018moleculenet}} & &  \\
&
  \multicolumn{1}{c|}{HIV \citep{wu2018moleculenet}} &  &  \\
\multicolumn{1}{|c|}{} &
  \multicolumn{1}{c|}{BACE \citep{wu2018moleculenet}} &  &  \\
\multicolumn{1}{|c|}{} &
  \multicolumn{1}{c|}{QM9 \citep{fang2023mol}} & \href{https://github.com/zjunlp/Mol-Instructions}{link} & random \\ \hline
Molecular generation &
  \multicolumn{1}{c|}{MOSES \citep{polykovskiy2018entangled}} &
  \multicolumn{1}{c|}{\href{https://github.com/molecularsets/moses}{link}} &
  predefined \\ \hline
Forward Reaction Prediction &
  \multicolumn{1}{c|}{\multirow{3}{*}{Mol-Instructions \citep{fang2023mol}}} &
  \multicolumn{1}{c|}{\multirow{3}{*}{\href{https://github.com/zjunlp/Mol-Instructions}{link}}} &
  \multirow{3}{*}{random} \\
Reagent Prediction &
  \multicolumn{1}{c|}{} &
  \multicolumn{1}{c|}{} &
   \\
Retrosynthesis &
  \multicolumn{1}{c|}{} &
  \multicolumn{1}{c|}{} &
   \\ \hline
Description-guided molecule design &
    \multicolumn{1}{c|}{\multirow{2}{*}{Mol-Instructions \citep{fang2023mol}}}
    &
    \multicolumn{1}{c|}{\multirow{2}{*}{\href{https://github.com/zjunlp/Mol-Instructions}{link}}}
    &
    \multirow{2}{*}{random} \\
Molecular description generation &  &
   & 
   \\ \hline
\end{tabular}
\end{table}

Fig. \ref{fig:model} shows our model and prompt format. Details on train/test splits are presented in Table \ref{tab:datalinks}. Datasets' descriptions with example instances are reported in Supplementary Information, Sec. 2.  
%Sec. \ref{sec:datasets}.

Given the presence of textual and molecular modalities, different tokenization technique is a crucial aspect of dataset design. One way to represent molecular structures is a simplified molecular-input line-entry system (SMILES) string \citep{weininger88smiles}. SMILES describe a molecule as a sequence of atoms in a depth-first traversal order and uses special symbols to depict branching, cycle opening/closing, bond types, and stereochemistry. We use the following tokenization:
\begin{itemize}
\item Textual domain sub-word tokens adopted from FLAN-T5 \citep{chung2022scaling} for natural language sequences; % 
\item Tokenization for SMILES: we annotate each SMILES token with special symbols: {\tt <sm\_\{token\}>} and extend the vocabulary with such tokens.
\end{itemize}

%Pre-training and fine-tuning \ourmodel{} large model required 10 days on 64 A100 80G GPUs from 8 NVIDIA DGX notes using the NVIDIA cluster architecture. Details are presented in Sec. \ref{sec:methods}.  

\subsection{Model and Training Configuration}

In our study, we predominantly employ a model featuring the default T5 architecture, which is derived from \citet{raffel2020exploring}. Our experimentation involves two model sizes: a base model consisting of 250 million parameters, characterized by 12 layers, a hidden state of 768 dimensions, a feed-forward hidden state of 3072 dimensions, and 12 attention heads; and a larger model with 780 million parameters, consisting of 24 layers, a hidden state of 1024 dimensions, a feed-forward hidden state of 4096 dimensions, and 16 attention heads.

For both models, we conduct pre-training with a language modeling (LM) objective and subsequent fine-tuning. The base models were trained using NVIDIA A4000 and A5000 GPUs, while the larger models were trained on NVIDIA's DGX cloud platform. Both the pre-training and fine-tuning stages were executed using the subsequent hyperparameters: a batch size of 1024, a learning rate set to 1e-4, and a weight decay of 0.01. The pre-training stage lasted for a single epoch, whereas the fine-tuning stage for 10 epochs. 

To execute the pre-training phase of our model with the LM objective, we leveraged two textual data sources in addition to one chemical data source. These textual data sources encompassed abstract texts extracted from Pubmed and patent descriptions derived from USPTO. All the textual data underwent a filtering process, eliminating documents that were not related to the chemistry domain. Consequently, the number of documents was curtailed to 13M for abstracts and 119K for patents. The chemical data component was sourced from the ZINC dataset, encompassing approximately 100 million documents. In aggregate, the textual data set contained 355M tokens for abstracts and 2.9B tokens for patents, whereas the chemical data encompassed 4.7B tokens.

The entirety of the investigations in this paper was conducted using the multi-task model, with the exception of the ablation part. Each multi-task model underwent fine-tuning by leveraging the entire spectrum of available datasets, encompassing all domains, as elucidated in Sec. \ref{tab:datalinks}. For data mixing and balancing we followed the “Examples-proportional mixing strategy” from \citet{raffel2020exploring}. The outcomes of these models are explicitly detailed in Sec. \ref{tab:mainres}. Conversely, in the context of ablation studies, fine-tuning was specifically performed utilizing only those datasets relevant to the corresponding domain, as detailed in the discussion.

\subsection{Nemo, Parallel Training, NVIDIA Cluster}

The training was performed using NVIDIA NeMo Toolkit \citep{Harper_NeMo_a_toolkit}, which consists of pre-built modules for end-to-end workflows in Automatic 
Speech Recognition (ASR), NLP, and Text-to-Speech (TTS) synthesis. NeMo uses PyTorch Lightning for optimized multi-node/multi-GPU (MNMG) mixed-precision training. In this work, we leveraged the NeMo NLP collection to train and evaluate our LMs. 
We trained our model on a variety of tasks such as information extraction, question answering, molecular property prediction, and description-guided molecule design using the NeMo toolkit.  A custom connector was added to extend the vocabulary size of the pre-trained model when continuing the training of the model with chemistry and biomedical datasets. The original vocabulary was extended to match the target vocabulary which was larger. The corresponding embedding matrix was initialized with learned embeddings of the original model. The extra tokens were initialized by re-using the first embeddings.

Data was parsed using Mem-Map Datasets from the NeMo toolkit to allow efficient data handling. The mem-map dataset relies on memory mapping directly to files, allowing the handling of very large datasets with small memory footprints and optimal reading speed. The data was loaded as raw text files and the tokenization occurred on-the-fly. Pre-fetching of the data mitigated the effects of online tokenization when compared to pre-tokenized data. The model was trained using tensor and pipeline parallelism \citep{10.1145/3458817.3476209}, both of which are model parallel methods for distributed training and are implemented in the NeMo toolkit for efficient scaling of large language model training. 
\section{Results and discussion}

\subsection{Use case: End-to-end drug discovery}

\begin{figure}
    \centering
    \includegraphics[width=0.8\linewidth]{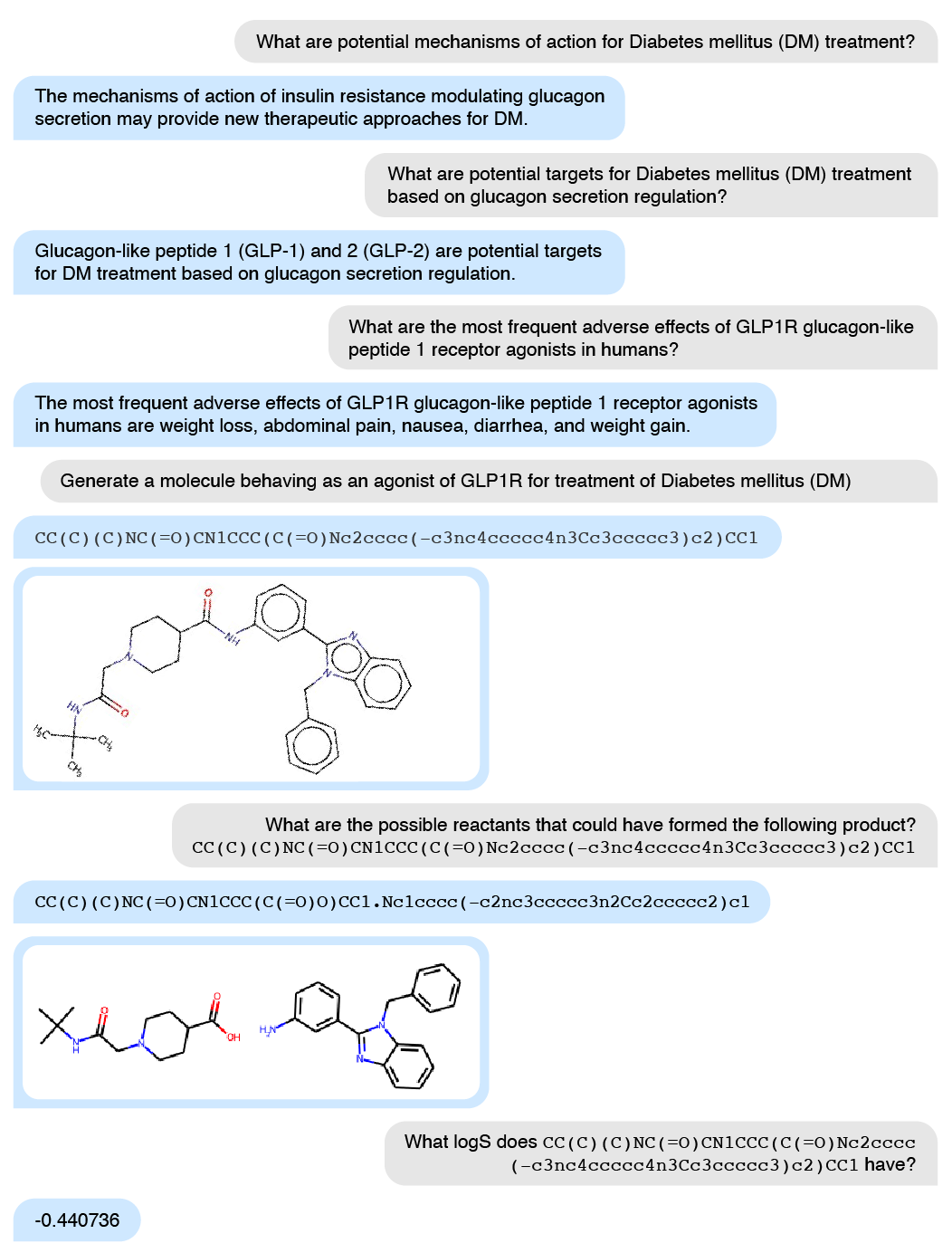}
    \caption{Input request from a human (gray color) and \ourmodel's response (blue color). }
    \label{fig:chat}
\end{figure}

In the first case study, we generate molecular structures against Diabetes mellitus (DM) using just one model, \ourmodel: discover biological targets with potential therapeutic activity, analyze the mechanism of action, generate molecular structure, propose one-step synthesis, and predict molecular properties. In a series of questions, we generate the model's responses using top-p sampling with values from 0.3 to 0.7 and step equals 0.05 and ask an expert chemist to pick the best response (Fig. \ref{fig:chat}). In total, we generate 200 SMILES on the molecule generation prompt and select one structure, {\tt CC(C)(C)NC(=O)CN1CCC(C(=O)Nc2cccc(-c3nc4ccccc4n3Cc3cc ccc3)c2)CC1}, as the most promising based on a chemical expert knowledge perspective. This semi-automated approach is efficient for discovering novel molecules and assessing their properties. We predict that further iterations of this model will require less supervision, and medicinal chemists will start using it as a side-car for generating and validating ideas.

%Table \ref{tab:line} shows example input and output from \ourmodel{}.

%\begin{table}[ht!]
%\caption{Example input and output from \ourmodel{} about R547 and Trifluridine drugs and a HCT-15 colorectal adenocarcinoma cell line.} \label{tab:line}
%\begin{tabular}{@{}|p{5cm}|p{7cm}|@{}}
%\hline
%Input request & \ourmodel's response \\ \hline
%What is R547? & X \\ %\hline
%Is R547 effective against HCT-15 colorectal adenocarcinoma cells, with IC50 values of below 1 micromolar? & R547 is highly effective against HCT-15 colorectal adenocarcinoma cells, with IC50 values of below 1 micromolar. \\ %\hline
%What is Trifluridine? & X \\ %\hline
%Is Trifluridine effective against HCT-15 colorectal adenocarcinoma cells, with IC50 values of below %1 micromolar? & Efficacy against HCT-15 colorectal adenocarcinoma cells is low. \\ 
%\hline
%\end{tabular}
%\end{table}

\subsection{Use case: Chemistry42 generative model}
\begin{table*}[ht!]
\centering
\caption{Comparison between \ourmodel{} and Chemistry42 models on JAK3 inhibitors generation. \ourmodel{} can discover multiple molecules passing all constraints, even though it only uses implicit knowledge about the protein target. Discovery rate (percentage of good molecules from all generated molecules) indicates that our models acts better than random combinatorial generator when solving the problem.} \label{tab:chemistry42}
\begin{tabular}{@{}|p{2.5cm}|p{2.5cm}|p{2.5cm}|p{2.5cm}|@{}}
\hline
& {\bf Combinatorial generator} &  {\bf \ourmodel{}} & {\bf Chemistry42} \\ \hline
Time & 24 hours & 45 minutes & 72 hours \\ \hline
Total molecules  & 73,000 & 7,200 & 382,000 \\ \hline
Good molecules & 30 & 8 & 5,841 \\ \hline
Discovery rate & 0.04\% & 0.11\% & 1.53\% \\ \hline

Best molecule & \includegraphics[width=0.8\linewidth]{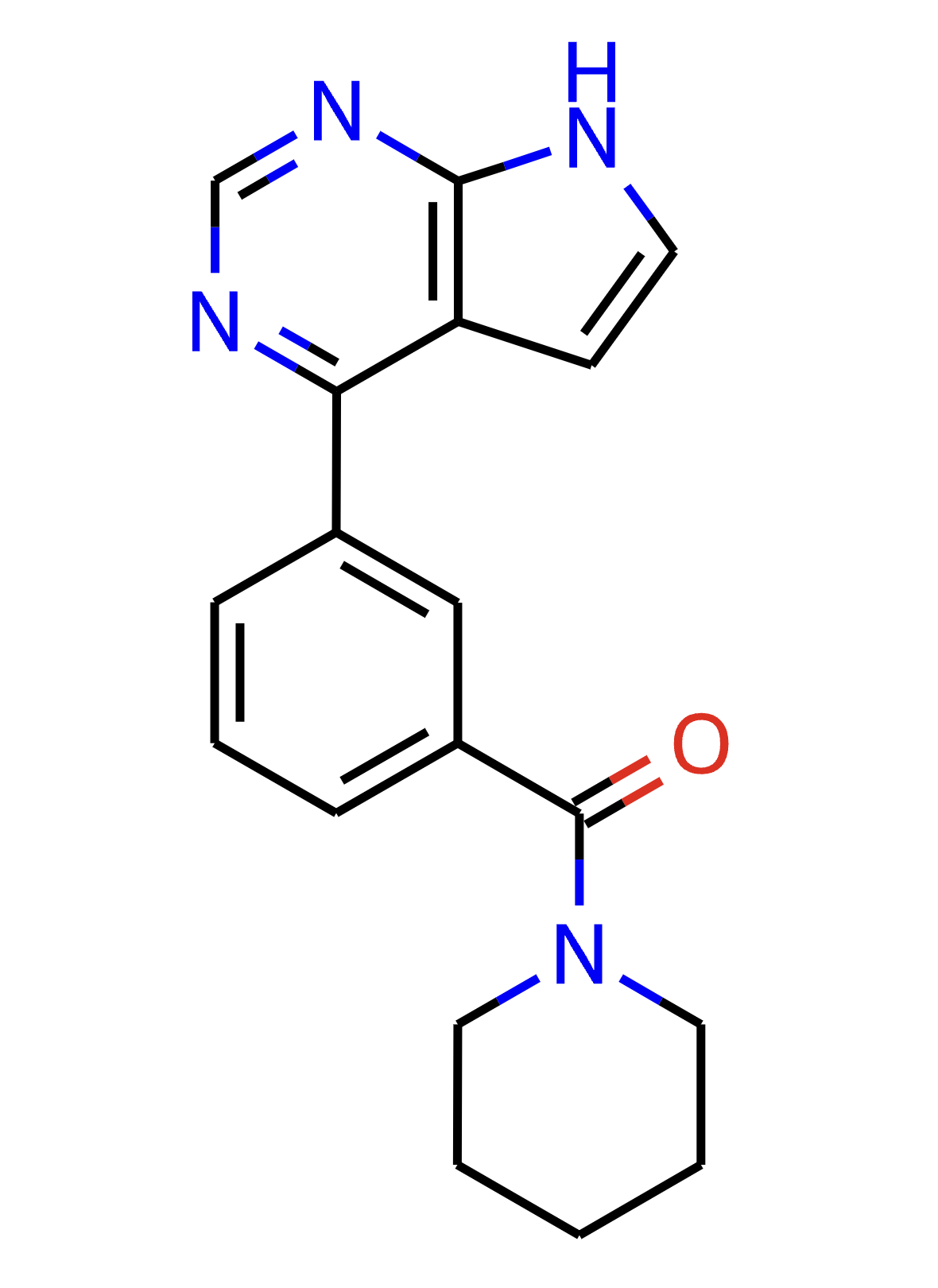} & \includegraphics[width=0.9\linewidth]{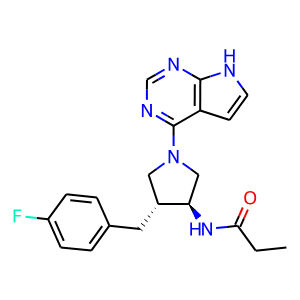} & \includegraphics[width=0.9\linewidth]{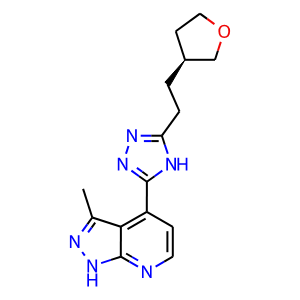} \\ \hline
\end{tabular}
\end{table*}

Chemistry42 is Insilico Medicine's AI drug discovery platform that efficiently generates novel active molecules using $42$ generative models \citep{chemistry42}. In this experiment, we apply  \ourmodel{} to one of the published case study setups available on demand at \href{demo.chemistry42.com}{demo.chemistry42.com}---Structure-Based Design of Janus Kinase 3 Inhibitors. In Chemistry42, we use 3LXK crystal structure, pharmacophore hypothesis, and a set of physicochemical properties to set up the search space for the generative models. All generative models search the chemical space to find the best possible structures.

Chemistry42 provides a set of filters ans reward modules. The 2D modules comprise of various tools including Medicinal Chemistry Filters (MCFs), Lipinski’s Rule of Five (Ro5), and descriptors for Drug-likeness, Weighted atom-type portion, Drug-likeness and Novelty, the synthetic accessibility (SA) scores. Additionally, Chemistry42 use the Self-Organizing Maps (SOM) Classifier Module to navigate the generation of molecular structures towards a specific target class in the chemical space. The Structure Morphing module, another integral part of 2D modules, is utilized to tackle metabolic instability issues.

The 3D modules include the ConfGen Module, which is responsible for generating conformational ensembles for each molecular structure. Subsequently, these molecules are ranked based on their intrinsic rigidity using a flexibility assessment tool. The 3D similarity between the generated structures and a reference molecule is evaluated using the 3D-Descriptors Module. The Pharmacophore Module is then used to find any matches with the specified pharmacophore hypothesis. The Shape Similarity Module plays its part in evaluating the 3D shape similarity to a reference molecule. Lastly, the Pocket Module and the Pocket-Ligand Interaction (PLI) modules are used to assess how well the molecules fit the chosen binding site.

In this experiment, we replaced all 42 generative models with \ourmodel{} and generated a set of structures using a prompt ``Generate a random druglike small inhibitor molecule for the Janus Kinase 3 JAK3 that contains a classic kinase hinge binding motif''. Note that \ourmodel{} does not have access to the specific crystal structure and other required properties, so the model generated molecules using solely its knowledge about JAK3.

In Tab. \ref{tab:chemistry42}, we compare generation results using a combinatorial generator \citep{polykovskiy2020molecular}, Chemistry42 \citep{chemistry42}, and our model. In just $45$ minutes (consisting of $15$ minutes for generation and $30$ minutes for scoring in Chemistry42), our model discovered $8$ molecules satisfying all the 2D and 3D requirements; see \citet{chemistry42} for more details on requirements. All these structures have a hinge binder and properly bind in the active site. While our model can discover multiple molecules satisfying all constraints, the discovered structures are currently worse than those found in $72$ hour generations in Chemistry42, since \ourmodel{} does not yet learn from the reinforcement learning feedback during generation and because it does not have exact knowledge of the experiment setup. In future work, we will expand our model with reinforcement learning capabilities to improve generation quality.

%\subsection{Use case: Life Star 1 Robotics lab}
%In the second case study, we validated the \ourmodel's predictions using Insilico Medicine's {\bf Life Star 1} Robotics lab. The goal was to predict antiproliferation activity of $10$ drugs against $8$ different colorectal adenocarcinoma cell lines. Robotics lab measured the activity of $36$ drug/cell line pairs, and the goal for the model was to predict which pairs were active. The full dataset will be published in a separate robotics lab paper. Using this data, we ask \ourmodel{} to generate responses to the prompt ``is {\tt <DRUG>} effective against {\tt <CELL\_LINE>} colorectal adenocarcinoma cells, with IC50 values of below 1 micromole?''. For each pair, we run the model $50$ times and average the predictions to estimate the activity probability. The model correctly identified $27$ pairs, while the majority vote baseline only correctly identified $19$.  

\subsection{Comparison of multi-task models}

Table \ref{tab:mainres} compares \ourmodel{} base and large models with two existing NLP encoder-decoder models (general-domain FLAN \citep{chung2022scaling} and domain-specific SciFive \citep{phan2021scifive}), and a multi-domain encoder-decoder model MolT5 \citep{edwards2022translation}. The table contains metrics for each task and model, with the results of the top-performing base model emphasized in bold. First, FLAN base and \ourmodel{} base exhibit similar results on NLP tasks on average, demonstrating superior performance on different tasks. With single-domain models for tasks such as NER or NLI, where molecule information is not required, traditional LMs may indeed provide the best results. However, when it comes to molecular tasks that involve molecular data, \ourmodel{} has distinct advantages over similar-scale models due to its specialized architecture and ability to effectively incorporate and process molecule-related information. In particular, \ourmodel{} benefits from training on diverse datasets and the proposed tokenization approach, outperforming baselines (including FLAN) with a significant gap in molecular tasks. For regression tasks, nach0 shows the best results on both RMSE and R2 scores. Moreover, in the molecular generation task, nach0 substantially surpasses FLAN by the FCD metric, which assesses the closeness of the generated molecules distribution to the ground truth. We added this explanation to the manuscript.
Second, as expected, large \ourmodel{} performed best among all the models. In terms of base models, \ourmodel{} base achieved the best results on chemical and cross-domain tasks over existing models, confirming that pre-training on two types of data with different tokens can be effective. 
%chemical-related encoder-decoder model (T5Chem \citep{lu2022unified})

Furthermore, we conducted zero-shot experiments involving \ourmodel{}, FLAN, and SciFive (all base versions) in an information retrieval task. The objective was to detect whether an abstract is relevant to a given disease or gene query. The dataset used for these experiments, along with its specific details, can be found in \citet{tutubalina2022comprehensive}. In these experiments, we employed the following prompt: ``Given the following passage, answer the question: Is the following text related to the \textit{synonym}? Passage: \textit{text}''. To evaluate the models' performance, we utilized precision (P), recall (R), and F-measure (F1). Our findings indicate that \ourmodel{} achieved an F1 score of 82.24\% (with a recall of 96.32\% and precision of 71.76\%), while FLAN and SciFive achieved F1 scores of 82.24\% and 77.20\%, respectively. However, it is worth noting that the supervised BERT-based pipeline from \citet{tutubalina2022comprehensive} achieved a higher F1 score of 88.81\%. Based on these results, we can conclude that these models exhibit the ability to perform slightly different NLP tasks in a zero-shot setup. However, they still fall significantly behind supervised models in terms of performance.

%We also provide metrics for molecular task. We show that \ourmodel{} benefits from training on diverse datasets and proposed tokenization approach, outperforming baselines with a significant gap.  For regression tasks, we present both RMSE and R2 scores. RMSE helps us see the direct prediction error, while R2, which handles outliers better, shows how well our predictions correlate the actual outcomes. 

\begin{table*}[t!]
\small
\centering
\caption{Full results of \ourmodel{} on NLP, CHEM and cross-domain tasks in comparison with FLAN (250M parameters), SciFive (220M parameters), MolT5 (220M parameters). All models are trained in a multi-task fashion. Bold number is the highest score on each dataset and the underscore stands for the second best result over \textbf{base models only}. We mark the results of Nach0 Large with a \textcolor{ForestGreen}{green color} to indicate improvements over Nach0 Base.
}\label{tab:mainres}%
\setlength{\tabcolsep}{3pt}
\begin{tabular}{@{}|p{5cm}|l|p{1.2cm}|p{1.2cm}|p{1.2cm}|p{1.4cm}||p{1.4cm}|@{}}
\hline
\multirow{2}{*}{Dataset} & \multirow{2}{*}{Metric}  & \textbf{MolT5} & \textbf{SciFive} & \textbf{FLAN}& \multicolumn{2}{c|}{\textbf{\ourmodel{}}} \\\cline{3-7}
 &  &  \multicolumn{4}{c||}{Base} & Large  \\\hline

%Named Entity Recognition &  & 68.19\% & 80.01\% & 76.19\% & 80.35\% & 86.27\% \\
BC5-chem & \multirow{5}{*}{F-1$\uparrow$} & 77.82\% & \textbf{91.02\%} & 88.03\% & \underline{90.96\%} & \textcolor{ForestGreen}{92.78\%} \\
BC5-disease &  & 71.62\% & \textbf{82.24\%} & 78.29\% & \underline{81.67\%} & \textcolor{ForestGreen}{85.51\%} \\
NCBI-disease &  & 74.96\% & \underline{84.22\%} & 81.37\% & \textbf{84.30\%} & \textcolor{ForestGreen}{85.82\%} \\
BC2GM &  & 53.47\% & \underline{69.55\%} & 62.53\% & \textbf{71.12\%} & \textcolor{ForestGreen}{80.41\%} \\
JNLPBA &  & 63.06\% & \underline{72.99\%} & 70.74\% & \textbf{73.70\%} & \textcolor{ForestGreen}{79.80\%} \\ \hline
%PICO extraction &  & 67.37\% & 67.32\% & 69.48\% & 67.60\% & 73.69\% \\
EBM PICO & F1$\uparrow$ & 67.37\% & 67.32\% & \textbf{69.48\%} & \underline{67.60\%} &  \textcolor{ForestGreen}{94.44\%} \\ \hline
%Textual Entailment &  & 57.61\% & 75.51\% & 85.17\% & 78.76\% & 91.73\% \\
MedNLI & \multirow{2}{*}{Accuracy$\uparrow$} & 58.69\% & 70.29\% & \textbf{79.66\%} & \underline{73.40\%} & \textcolor{ForestGreen}{89.22\%} \\
SciTail &  & 56.54\% & 80.73\% & \textbf{90.68\%} & \underline{84.12\%} & \textcolor{ForestGreen}{93.87\%} \\ \hline
%Relation Extraction &  & 59.54\% & 66.63\% & 79.09\% & 82.59\% & 88.54\% \\
ChemProt & \multirow{3}{*}{F-1$\uparrow$} & 70.52\% & 75.83\% & \textbf{84.38\%} & \underline{83.61\%} & \textcolor{ForestGreen}{94.46\%} \\
DDI &  & 56.02\% & 59.53\% & \underline{85.96\%} & \textbf{88.69\%} & \textcolor{ForestGreen}{93.13\%} \\
GAD &  & 52.10\% & 64.53\% & \underline{66.93\%} & \textbf{75.47\%} & \textcolor{ForestGreen}{78.24\%} \\ \hline
%Sentence similarity &  & 24.55\% & 56.51\% & 61.21\% & 52.58\% & 75.58\% \\
BIOSSES & Pearson$\uparrow$ & 24.55\% & \underline{56.51\%} & \textbf{61.21\%} & 52.58\% & 52.37\% \\ \hline
%Document Classification &  & 70.24\% & 72.49\% & 72.37\% & 80.40\% & 85.23\% \\
HoC & F-1$\uparrow$ & 70.24\% & \underline{72.49\%} & 72.37\% & \textbf{80.40\%} & \textcolor{ForestGreen}{85.86\%} \\ \hline
%Question answering (Yes/No) &  & 55.42\% & 69.86\% & 74.97\% & 69.09\% & 77.39\% \\
PubMedQA & \multirow{2}{*}{F-1$\uparrow$} & 49.12\% & \underline{59.44\%} & \textbf{62.80\%} & 58.76\% & \textcolor{ForestGreen}{74.21\%} \\
BioASQ &  & 61.71\% & \underline{80.29\%} & \textbf{87.14\%} & 79.43\% & \textcolor{ForestGreen}{89.21\%} \\ \hline
%Question answering (Multi Choice) &  & 25.97\% & 25.06\% & 25.42\% & 26.61\% & 37.79\% \\
MedMCQA and MMLU & Accuracy$\uparrow$ & \underline{25.97\%} & 25.06\% & 25.42\% & \textbf{26.61\%} & \textcolor{ForestGreen}{46.10\%} \\ \hline
%Question answering (Open) &  & 4.52\% & 5.83\% & 5.10\% & 6.30\% & 6.62\% \\
MedMCQA-Open & BLEU-2$\uparrow$ & 4.52\% & \underline{5.83\%} & 5.10\% & \textbf{6.30\%} & 2.26\% \\ \hline
%Reaction &  &  &  &  &  &  \\
Reagent prediction & Accuracy@top1$\uparrow$ & 1.10\% & 3.80\% & \underline{4.00\%} & \textbf{6.30\%} & \textcolor{ForestGreen}{13.08\%} \\
Retrosynthesis & Accuracy@top1$\uparrow$ & 15.00\% & \underline{31.00\%} & \underline{31.00\%} & \textbf{53.00\%} & \textcolor{ForestGreen}{56.26\%} \\
Forward reaction prediction & Accuracy@top1$\uparrow$ & 27.00\% & \underline{60.00\%} & 59.00\% & \textbf{88.00\%} & \textcolor{ForestGreen}{89.94\%} \\ \hline 
%Property &  &  &  &  &  &  \\ 
BACE & BA$\uparrow$ & 0.58 & \underline{0.65} & \underline{0.65} & \textbf{0.74} & 0.71  \\
BBBP & BA$\uparrow$ & 0.55 & \underline{0.66} & 0.6 & \textbf{0.67} & \textcolor{ForestGreen}{0.68} \\
HIV & BA$\uparrow$ & 0.5 & \underline{0.53} & \underline{0.53} & \textbf{0.56} & \textcolor{ForestGreen}{0.60} \\
\multirow{2}{*}{HFE} & R2$\uparrow$ & -0.36 & 0.51 & 0.55 & \textbf{0.77} & \textcolor{ForestGreen}{0.78} \\
 & RMSE$\downarrow$ & 1.1 & \underline{0.4} & 0.37 & \textbf{0.19} & {0.19} \\
\multirow{2}{*}{HOMO-LUMO} & R2$\uparrow$ & 0.98 & \underline{0.99} & \underline{0.99} & \textbf{1.00} & {1.00} \\
 & RMSE$\downarrow$ & 0.0008 & \underline{0.0003} & \underline{0.0003} & \textbf{0.0001} & {0.0001} \\
\multirow{2}{*}{LOGD} & R2$\uparrow$ & -0.6 & \underline{-0.27} & -0.32 & \textbf{0.28} & {0.28} \\
 & RMSE$\downarrow$ & 2.4 & \underline{1.9} & \underline{1.9} & \textbf{1.1} & {1.1} \\
\multirow{2}{*}{LOGS} & R2$\uparrow$ & -0.49 & \underline{0.31} & 0.001 & \textbf{0.48} & {0.48} \\
 & RMSE$\downarrow$ & 1.4 & \underline{0.63} & 0.91 & \textbf{0.48} & {0.48} \\ \hline
%Molecular generation &  &  &  &  &  &  \\
\multirow{9}{*}{MOSES} & Valid$\uparrow$ & 
\underline{98.30\%} & 95.79\% & 97.63\% & \textbf{99.86\%} & {99.93\%} \\
 & Unique@10000$\uparrow$ & 99.93\% & 99.94\% & \textbf{99.95\%} & \underline{99.92\%} & \textcolor{ForestGreen}{99.97\%} \\
 & FCD/Test$\downarrow$ & \underline{0.5212} & 0.5778 & 0.5289 & \textbf{0.3106} & \textcolor{ForestGreen}{0.3038} \\
 & SNN/Test$\uparrow$  & \underline{0.5745} & 0.5688 & 0.5742 & \textbf{0.6118} & \textcolor{ForestGreen}{0.6222} \\
 & Frag/Test$\uparrow$  & \underline{0.9974} & 0.9967 & 0.9965 & \textbf{0.9985} & \textcolor{ForestGreen}{1.00} \\
 & Scaf/Test$\uparrow$  & 0.8748 & 0.8737 & \underline{0.8823} & \textbf{0.9205} & \textcolor{ForestGreen}{0.9292} \\
 & IntDiv$\uparrow$  & 0.8460 & \underline{0.8464} & 0.8462 & \textbf{0.8478} & \textcolor{ForestGreen}{0.8585} \\
 & Filters$\uparrow$  & \underline{98.89\%} & 98.67\% & 98.68\% & \textbf{99.54\%} & \textcolor{ForestGreen}{99.67\%} \\
 & Novelty$\uparrow$  & \underline{93.92\%} & \textbf{93.98\%} & 93.67\% & 87.60\% & \textcolor{ForestGreen}{93.87\%} \\ \hline
%Cross-domain &  &  &  &  &  &  \\
Description-guided molecule design &  BLEU-2$\uparrow$  & 30.32\% & \underline{44.17\%} & 43.64\% & \textbf{48.97\%} & {48.76\%} \\
Molecular description generation &  BLEU-2$\uparrow$  & 35.61\% & \underline{39.56\%} & 38.58\% & \textbf{43.91\%} & {41.73\%} \\ \hline
\end{tabular}
\end{table*}

\subsection{Ablations}
To examine the impact of cross-domain data on multi-task fine-tuning, we conducted training on mono-domain data. The results of four pre-trained checkpoints (SciFive, FLAN, MolT5, \ourmodel) fine-tuned exclusively on NLP data are presented in Supplementary Information, Sec. 1. When considering average performance on the NLP group, \ourmodel, SciFive, and FLAN exhibit similar results, MolT5 achieves lower scores compared to the other models. 

Next, we investigate how  chemical tasks groups combination effects on joint model performance in comparison with individual models trained on each separate chemical tasks group---on predictive tasks group, on reaction tasks group and molecular generation/cross-domain tasks group. We perform the same experiments with MolT5 model to elaborate on how pretraining data and special chemical tokens affect the quality of the model on chemical tasks. 

The results of this ablation study can be found in Tab. \ref{tab:chem_tasks_abl} and show that \ourmodel{} benefits from combining chemical tasks group---model trained on the whole set of chemical data without NLP outperforms in total set of metrics models trained on distinct task groups. It is important to mention that despite the joint model showing worse metrics than the model trained only on molecular generation and cross-domain tasks, it works better since it does not overfit on training data---the novelty metric is more prevail here over all other molecule generation metrics.

Also, experiments show that the special chemical tokens and pre-training on both natural language and chemical data improve the model quality---\ourmodel{} outperforms MolT5 baseline or show equal metrics on each chemical task group. We miss some MolT5 metrics on molecule generation task since it produces non-valid SMILES sequences.

\begin{table*}[t!]
\small
\centering
\caption{Performance of \ourmodel{} on chemical tasks groups in comparison with MolT5. We list the scores for each task (see Supplementary Information about datasets and metrics). Bold number is the best result on each dataset. All models are base models.}\label{tab:chem_tasks_abl}
\setlength{\tabcolsep}{4pt}
\begin{tabular}{@{}|p{4.5cm}|l|l|lll|l|lll|@{}}
\hline
\multirow{2}{*}{\textbf{Dataset}} & \multirow{2}{*}{\textbf{Metric}} & \multicolumn{4}{c|}{\textbf{\ourmodel{}}}  & \multicolumn{4}{c|}{\textbf{MolT5}} \\ \cline{3-10} 
& & All & Pred. & React. & Mol. Gen. & All & Pred. & React. & Mol. Gen. \\
\hline
\rowcolor[HTML]{C0C0C0}
\multicolumn{10}{c}{Prediction tasks} \\
\hline
BACE & BA $\uparrow$ & \textbf{0.74} & 0.67 & - & - & 0.58 & 0.52 & - & - \\
\hline
BBBP & BA $\uparrow$ & \textbf{0.67} & 0.62 & - & - & 0.55 & 0.57 & - & - \\
\hline
HIV & BA $\uparrow$ & 0.56 & \textbf{0.65} & - & - & 0.5 & 0.51 & - & - \\
\hline
\multirow{2}{*}{HFE} & R2 $\uparrow$ & \textbf{0.77} & 0.015 & - & -  & -0.36 & -0.74 & - & - \\
 & RMSE $\downarrow$ & \textbf{0.19} & 0.81 & - & - & 1.1  &  1.4 & - & - \\
\hline
\multirow{2}{*}{HOMO-LUMO} & R2 $\uparrow$ & \textbf{1.0} & \textbf{1.0} & - & - & 0.98 &  0.94 & - & - \\
 & RMSE $\downarrow$ & 1e-4 & \textbf{1e-5} & - & - & 7e-4  & 2e-4 & - & - \\
 \hline
\multirow{2}{*}{LOGD} & R2 $\uparrow$ & \textbf{0.28} & 0.27 & - & - & -0.6 & -2.9 & - & - \\
 & RMSE $\downarrow$ & \textbf{1.1} & \textbf{1.1} & - & - & 2.4 & 5.7 & - & - \\
 \hline
\multirow{2}{*}{LOGS} & R2 $\uparrow$ & \textbf{0.48} & 0.32 & - & - & -0.49 & -1.2 & - & - \\
 & RMSE $\downarrow$ & \textbf{0.48} & 0.62 & - & - & 1.4 & 2.0 & - & - \\
 \hline
 \rowcolor[HTML]{C0C0C0}
 \multicolumn{10}{c}{Reaction tasks} \\
 \hline
 Reagent prediction & Accuracy $\uparrow$ & 0.063 & - & \textbf{0.14} & - & 0.011 & - & 0.13 & - \\
 \hline
 Retrosynthesis & Accuracy $\uparrow$ & \textbf{0.53} & - & 0.39 & - & 0.15 & - & 0.39 & - \\
 \hline
 Forward reaction prediction & Accuracy $\uparrow$ & 0.88 & - & \textbf{0.89} & - & 0.27 & - & \textbf{0.89} & - \\
 \hline
 \rowcolor[HTML]{C0C0C0}
 \multicolumn{10}{c}{Molecular generation and cross-domain tasks} \\
 \hline
 \multirow{9}{*}{Molecule generation} & Validity $\uparrow$ & 99.86\% & - & - & \textbf{99.99\%} & 98.3\% & - & - & 0.0\% \\
 & Unique@10000 $\uparrow$ & 99.92\% & - & - & 99.81\% &  \textbf{99.93\%} & - & - & N/A \\
 & FCD/Test $\downarrow$ & 0.3106 & - & - & \textbf{0.2411} & 0.5212 & - & - & N/A \\
 & SNN/Test $\uparrow$ & 0.6118 & - & - & \textbf{0.6551} & 0.5745 & - & - & N/A \\
 & Frag/Test $\uparrow$ & 0.9985 & - & - & \textbf{0.9988} & 0.9974 & - & - & N/A \\
 & Scaf/Test $\uparrow$ & 0.9205 & - & - & \textbf{0.9403} & 0.8748 & - & - & N/A \\
 & IntDiv $\uparrow$ & 0.8478 & - & - & \textbf{0.8493} & 0.846 & - & - & N/A \\
 & Filters $\uparrow$ & 99.54\% & - & - & \textbf{99.95\%} & 98.89\% & - & - & N/A \\
 & Novelty $\uparrow$ & 87.6\% & - & - & 64.34\% & \textbf{93.92\%} & - & - & N/A \\
 \hline
 Description-guided molecule gen. & BLEU-2 $\uparrow$ & 48.97\% & - & - & \textbf{52.90\%} & 30.32\% & - & - & 30.78\% \\
 \hline
 Molecular description generation & BLEU-2 $\uparrow$ & 43.91\% & - & - & \textbf{46.22\%} & 35.61\% & - & - & 31.32\% \\
 \hline
\end{tabular}

\end{table*}

\subsection{Comparison with ChatGPT}

Recently, a comprehensive benchmark for biomedical text generation and mining problems with ChatGPT was conducted, revealing its poor performance on several biomedical NLP benchmark datasets  \citep{tang2023does,chen2023comprehensive}. \citet{chen2023comprehensive} specifically evaluated ChatGPT on a BLURB benchmark \citep{gu2021domain}, which encompasses BC5-chem, BC5-disease, NCBI-disease, BC2GM, JNLPBA, EMB-PICO, ChemProt, DDI, GAD, BIOSSES, HoC, PubMedQA, BioASQ. In particular, ChatGPT got an average BLURB score of 48.27 on NER, while fine-tuned BERT achieved 86.27. For more details on evaluation scores, please refer to \citet{chen2023comprehensive}.

In our evaluation setup, we focus on three specific datasets: EMB-PICO, MedMCQA-Open, and molecular description generation (Mol-Instructions). The inclusion of EMB-PICO dataset was driven by its practical importance. This dataset involves the task of identifying and extracting specific fragments of text related to the Population/Patient/Problem (P), Intervention (I), Comparator (C), and Outcome (O) elements from unstructured biomedical texts, such as research articles and clinical trial reports. It is worth noting that the clinical trial domain holds particular significance for inClinico, a transformer-based artificial intelligence software platform designed to predict the outcome of Phase II clinical trials \citep{ct2023}. The molecular generation task is relevant to the Chemistry42 platform \citep{chemistry42}.

To evaluate the zero-shot performance, we had to limit the evaluation to a subset of 2000 samples from the test set for each of the three datasets, considering the computational constraints of ChatGPT. As well we utilized the GPT-3.5-turbo model through the OpenAI API and multi-task \ourmodel{} base for evaluation purposes. In the case of the PICO dataset, ChatGPT achieved a word-level F1 score of 64.43\%, comparable to the results obtained by fine-tuned \ourmodel{} base on this subset (F1 score of 67.60\%). For MedMCQA-Open, ChatGPT achieved a BLEU2 score of 1.68\%, while the fine-tuned \ourmodel{} base attained a BLEU2 score of 6.30\%. In the molecular description generation task, ChatGPT achieved a BLEU2 score of 2.23\%, whereas the fine-tuned \ourmodel{} base excelled with a BLEU2 score of 42.80\%. Based on our preliminary findings, it is evident that utilizing ChatGPT directly leads to subpar performance compared to models trained specifically on the domain-specific dataset, how it was done in \ourmodel{}. 

\subsection{Discussion}

In this study, we pretrained and fine-tuned T5 models, which have an encoder-decoder architecture. Nevertheless, a broad range of model families, including T5, BERT-based BioMegatron \citep{shin2020biomegatron}, decoder-only PaLM \citep{chowdhery2022palm} and GPT \citep{brown2020language}, exist. To determine the most suitable architecture for pre-training and fine-tuning on chemical-related data, it may be necessary to evaluate these alternatives. We suggest it as a potential topic for future research.

There have been several efforts to train large language models (LLMs) on biomedical corpora, particularly on PubMed. Notable examples include BioGPT (347M and 1.5B) \citep{luo2022biogpt}, PubMedGPT (2.7B) \citep{PubMedGPT}, and Galactica (120B) \citep{taylor2022galactica}. Through our experiments with scaling from a base model (250M) to a large model (780M), we demonstrated the benefits of scale on several datasets. Based on our findings, we can conclude that scaling can further enhance the chemical capabilities of models, particularly in terms of generation and reasoning skills.

\subsubsection{Limitations}

\subsubsection*{Key LLM capabilities for chemistry}
Although our LM was able to reach state-of-the-art performance on several chemistry-related benchmarks, our human evaluations clearly suggested that these models are not at the chemist expert level. In order to bridge this gap, several new LLM capabilities need to be researched and developed including (i) knowledge alignment between textual and chemical sources as well as domain-specific knowledge graphs; (ii) ability to perform chemical reasoning and provide explanations for their predictions; (iii) ability to learn from and adapt to feedback from human experts, (iv) ability to generate novel chemical reactions and materials.

\subsubsection*{Molecular representations}
One limitation of our LM is its focus on string representations of molecules, specifically the SMILES notation. Although SMILES is a widely used notation for representing molecules, it provides only 2D information of the molecule, missing the 3D geometry and spatial arrangement of atoms and bonds in a molecule. This can result in inaccuracies in predicting molecular properties and interactions. To address these limitations, it would be beneficial to incorporate additional modalities of molecules, such as the molecular graphs in terms of 2D or 3D representations, in the training of the language model. 

Another significant drawback of the SMILES format is the absence of a one-to-one translation between molecules and SMILES strings. Typically, a molecule can have multiple SMILES representations that differ from each other due to factors such as the starting atom, molecular graph traversal, and kekulization. In practice, SMILES strings are often converted to a canonical form using an unambiguous algorithm. A molecular representation called SELFIES \citep{krenn2020self,krenn2022selfies} was defined from scratch to be attractive as a sequential representation for molecules. All random SELFIES are valid molecular representations. SELFIES was extened to treat molecular groups as well \citep{cheng2023group}. As SELFIES have been repeatedly shown to have advantages over other representations in the context of generative models, exploring their use as the main representation for a language model is a future potential direction.

\subsubsection*{Prompt design}
Our language model has a limitation in that it heavily relies on the quality and specificity of the prompts, as well as the potential for biases in both the training data and the prompts themselves. To enhance the performance of the model, incorporating domain-specific and information-rich prompts is essential. One potential approach to achieving this is by leveraging the knowledge of domain experts to design effective biomedical prompts. Yet, over-reliance on domain-specific prompts may lead to a lack of diversity in the model's responses, which can limit its usefulness.

\subsubsection*{Chemical diversity}
Mol-Instructions includes cross-domain datasets that consist of compounds and their corresponding descriptions collected from PubChem. PubChem is a publicly available database administered by the National Center for Biotechnology Information (NCBI). It is important to note that the datasets primarily encompass current drugs and known chemical probes, representing only a fraction of the vast predicted chemical space. Furthermore, these datasets do not encompass testing on novel chemical diversity distinct from molecules documented in the literature.

\section{Conclusion}\label{sec13}
Our study integrates a diverse range of one-domain and multi-domain task types and biomolecular text instructions to address the landscape of chemical research on drug design, reaction prediction, and retrosynthesis and leverage the advancements in NLP and LLMs. EThe multi-domain training approach allows our model, nach0, to leverage a broader understanding of both chemical and linguistic knowledge. xtensive experiments and two case studies demonstrate that \ourmodel's capabilities in translating between natural language and chemical language enable it to tackle tasks effectively.  Considering the unique training methodology and the broader scope of tasks that our model can effectively handle, we believe our work presents a significant contribution to the field.

Based on our findings, we foresee several promising directions for future research. One direction could involve 
such as protein sequences, which would require adding special tokens into the model similar to SMILES. 
This task could be easily achieved with Group SELFIES.
New modalities require collecting diverse tasks with natural language prompts for fine-tuning.
A second direction involves extending NLP datasets and conducting zero-shot evaluations to assess the reasoning and generalization capabilities of \ourmodel.
Finally, exploring the fusion of information from textual sequences and relevant knowledge graphs as input in a self-supervised approach remains an area to be explored.

\section*{Author Contributions}
These authors contributed equally: Micha Livne, Zulfat Miftahutdinov, Elena Tutubalina, Maksim Kuznetsov.

ET, DP, AA, and AZ contributed to the conception and design of the work.
ET, ZM, and MK contributed to the data acquisition and curation. 
ZM, MK, ML, AC, AB, and AJ contributed to the technical implementation with the NeMo framework, provided technical and infrastructure guidance.
ET, ZM, and MK contributed to the evaluation framework used in the study.  
All authors contributed to the drafting and revising of the manuscript.

\section*{Conflicts of interest}
The authors declare no competing interests.
This study is a collaboration of NVIDIA and Insilico Medicine employees.

\subsection*{Data availability}
All datasets used in the study for pre-training and fine-tuning are publicly available. 

\subsection*{Code availability}
The nach0 framework is available for research purposes:
\begin{itemize}
    \item nach0 Base is available via \url{https://huggingface.co/insilicomedicine/nach0\_base};
    \item nach0 Large is available via \url{https://huggingface.co/insilicomedicine/nach0\_large};
    \item For pre-processing scripts, see \url{https://github.com/insilicomedicine/nach0}.
\end{itemize}

%\section*{Acknowledgements}
%The Acknowledgements come at the end of an article after Conflicts of interest and before the Notes and references.

%%%END OF MAIN TEXT%%%

%The \balance command can be used to balance the columns on the final page if desired. It should be placed anywhere within the first column of the last page.
\section{Supplementary}

\subsection{NLP Ablation}
To examine the impact of cross-domain data on multi-task fine-tuning, we conducted training on mono-domain data. The results of four pre-trained checkpoints fine-tuned exclusively on NLP data are presented in Supplementary Information, Tab. \ref{tab:nlp_only}. Several noteworthy observations can be made based on these findings.

Firstly, when considering average performance, \ourmodel, SciFive, and FLAN exhibit similar results. However, each model demonstrates superior performance on different tasks. FLAN, being a general-domain model, outperforms others in textual entailment, binary QA, and sentence similarity. On the other hand, the domain-specific SciFive shows best results in NER, while \ourmodel{} -- in relation extraction, classification, and multi-choice QA.

Secondly, MolT5 achieves lower scores compared to the other models. This can be related to the pre-training strategy, where molecules and natural language texts share the same tokens in the semantic space. In contrast, \ourmodel{} utilizes specialized tokenization for molecular data, which does not significantly impact overall performance on NLP tasks compared to SciFive and FLAN.

\begin{table*}[hb!]
\centering
\caption{Performance of \ourmodel{} on NLP tasks in comparison with FLAN, SciFive, MolT5. We list the scores for each task (see Sec. \ref{sec:datasets} about datasets and metrics). All models are base models.}\label{tab:nlp_only}
\begin{tabular}{l|llll}
\hline
 & \ourmodel{}  & FLAN-T5    & SciFive     & MolT5\\
\hline
\rowcolor[HTML]{C0C0C0} 
Named Entity Recognition          & 80.63\%          & 75.01\%          & \textbf{81.14\%} & 56.48\% \\
BC5-chem                          & 91.14\%          & 87.56\%          & 91.81\%          & 64.28\% \\
BC5-disease                       & 81.72\%          & 76.61\%          & 82.33\%          & 61.56\% \\
NCBI-disease                      & 84.43\%          & 79.46\%          & 85.33\%          & 54.74\% \\
BC2GM                             & 72.44\%          & 61.75\%          & 72.76\%          & 45.87\% \\
JNLPBA                            & 73.42\%          & 69.68\%          & 73.45\%          & 55.93\% \\
\hline
\rowcolor[HTML]{C0C0C0} 
PICO extraction                   & 67.10\%          & \textbf{68.94\%} & 67.62\%          & 66.39\% \\
EBM PICO                          & 67.10\%          & 68.94\%          & 67.62\%          & 66.39\% \\
\hline
\rowcolor[HTML]{C0C0C0} 
Textual Entailment                & 86.03\%          & \textbf{87.53\%} & 86.96\%          & 55.63\% \\
MedNLI                            & 81.28\%          & 81.75\%          & 82.90\%          & 55.67\% \\
SciTail                           & 90.77\%          & 93.31\%          & 91.01\%          & 55.58\% \\
\hline
\rowcolor[HTML]{C0C0C0} 
Relation Extraction               & \textbf{84.06\%} & 73.84\%          & 73.22\%          & 63.38\% \\
ChemProt                          & 89.40\%          & 84.48\%          & 82.77\%          & 75.98\% \\
DDI                               & 89.67\%          & 72.85\%          & 66.08\%          & 63.23\% \\
GAD                               & 73.11\%          & 64.19\%          & 70.82\%          & 50.93\% \\
\hline
\rowcolor[HTML]{C0C0C0} 
Sentence similarity               & 27.45\%          & \textbf{32.78\%} & 1.17\%           & 14.95\% \\
BIOSSES                           & 27.45\%          & 32.78\%          & 1.17\%           & 14.95\% \\
\hline
\rowcolor[HTML]{C0C0C0} 
Document Classification           & \textbf{83.83\%} & 75.48\%          & 82.49\%          & 70.99\% \\
HoC                               & 83.83\%          & 75.48\%          & 82.49\%          & 70.99\% \\
\hline
\rowcolor[HTML]{C0C0C0} 
Question answering (Yes/No)       & 63.87\%          & \textbf{65.04\%} & 63.66\%          & 51.6\% \\
PubMedQA                          & 51.32\%          & 50.36\%          & 52.04\%          & 47.20\% \\
BioASQ                            & 76.43\%          & 79.71\%          & 75.29\%          & 56.00\% \\
\hline
\rowcolor[HTML]{C0C0C0} 
Question answering (Multi Choice) & \textbf{27.71\%} & 25.61\%          & 26.29\%          & 25.54\%\\
MedMCQA and MMLU                  & 27.71\%          & 25.61\%          & 26.29\%          & 25.54\% \\
\hline
\rowcolor[HTML]{C0C0C0} 
Question answering (Open)         & \textbf{2.43\%}  & 2.34\%           & 2.25\%           & 1.83\% \\
MedMCQA-Open                      & 2.43\%           & 2.34\%           & 2.25\%          & 1.83\%
\end{tabular}
\end{table*}

%\subsection{Datasets}

\subsection{Chemistry: Tasks and Datasets}

We've integrated several chemical domain tasks from widely-used benchmarks and datasets. It covers distribution match, molecular property prediction, reaction prediction and related problems. Where it's possible, we use the provided standard train/validation/test split procedures, otherwise, we employ the random data split. We choose this data preparation strategy to enable comparison with baseline models, however, we don't guarantee that one can't find chemical objects with similar structures in the different subsets.

\subsubsection{MOSES}

MOSES dataset \cite{polykovskiy2020molecular} is a benchmarking platform that provides a large dataset and set of metrics to compare generative models on an unconditional molecular generation task. The dataset provided by MOSES contains almost 2 million samples filtered by MCF, PAINS, and additional rules. The metrics set estimates the quality of the generative model from several points of view: validity of generated structures, molecular distribution matching quality, and the ability of the model to produce novel, diverse molecules.

%\subparagraph{Technical Details}

%\noindent
\textbf{Evaluation metric:} The MOSES benchmark provides established set metrics for assessing the ability of models to produce unique, diverse, valid molecules similar to ground-truth distribution. In our work, we adopt several metrics: uniqueness, validity, novelty, internal diversity, similarity to a nearestneighbor (SNN), 
fragment similarity, scaffold similarity and FCD \cite{preuer2018fcd}. We've generated 30000 new molecules to compute these metrics.

\noindent
\textbf{Example on molecular distribution matching:} 

\textit{input text with prompt}:  Generate random molecule from MOSES dataset.

\textit{output text}: {\tt CC1C2CCC(C2)C1CN(CCO)C(=O)c1ccc(Cl)cc1}.

\subsubsection{Mol-Instructions}

The recently published Mol-Instructions dataset \cite{fang2023mol} covers three significant modalities: molecule-oriented instructions, protein-oriented instructions, and biomolecular text instructions.  In our study, we specifically focus on the first subset, which is the most relevant and contains chemical tasks.

\noindent
\textbf{Example on descriptor-guided molecule generation:} 

\textit{input text with prompt}: Synthesize a molecule that matches the given characteristics: The molecule is the (R)-enantiomer of aceprometazine. It is an enantiomer of a (S)-aceprometazine.

\textit{output text}: {\tt CC(=O)c1cc2c(cc1)Sc1ccccc1N2C[C@@H](C)N(C)C}.

\noindent
\textbf{Example on molecular description generation:} 

\textit{input text with prompt}: What can you tell me about this molecule?: {\tt C(=C(Cl)Cl)(Cl)Cl}.

\textit{output text}: The molecule is a manufactured chemical that is widely used for dry cleaning of fabrics and for metal-degreasing. It is also used to make other chemicals and is used in some consumer products.

\noindent
\textbf{Example on forward reaction prediction:} 

\textit{input text with prompt}: With the provided reactants and reagents, propose a potential product:  {\tt CO.C[Si](C)(C)C\#Cc1ccc(C=O)cc1.ClCCl.O=C([O-])[O-].[K+].[K+]}.

\textit{output text}: {\tt C\#Cc1ccc(C=O)cc1}.

\noindent
\textbf{Example on reagent prediction:} 

\textit{input text with prompt}:  Please propose potential reagents that might have been utilized in the provided chemical reaction: 

{\tt COC(=O)c1c(F)cc(NC(=O)c2cc(C(C)C)c(C(C)C)s2)cc1F} $\gg$ 

{\tt CC(C)c1c(C(C)C)sc(C(=O)Nc2cc(F)c(C(=O)O)c(F)c2)c1}.

\textit{output text}: {\tt [Na+].[OH-]}

\noindent
\textbf{Example on retrosynthesis:} 

\textit{input text with prompt}:  Provide a list of potential reactants that may have produced the given product.: {\tt Cc1ccc(-c2ccccc2N)cc1}

\textit{output text}: {\tt Cc1ccc(B(O)O)cc1.Nc1ccccc1I}

\subsubsection{Property Prediction}

We adopt several binary classification and regression tasks from the MoleculeNet benchmark to assess the model's ability to predict molecular properties. 

%\subparagraph{Technical Details}
\textbf{Evaluation metric:} Binary classification tasks include BBBP, HIV, and BACE datasets from MoleculeNet \cite{wu2018moleculenet} and use balanced accuracy  as the main metric. Regression tasks involve ESOL, FreeSOLV and Lipo datasets from MoleculeNet \cite{wu2018moleculenet}, QM9 dataset from MolInstructions \cite{fang2023mol} and rely on the R2 metric. In our work, we utilized the code provided by the MoleculeNet benchmark to prepare data splits.

\noindent
\textbf{Example on the BBBP classification task:} 

\textit{input text with prompt}:  Can {\tt CN(C)[C@H]1[C@@H]2C[C@H]3C(=C(O)c4c(O)cccc4[C@@]3 (C)O)C(=O)[C@]2(O)C(=O)C(=C(/O)NCN5CCCC5)C1=O} penetrate the BBB?

\textit{output text}: 1

\noindent
\textbf{Example on HIV classification task:} 

\textit{input text}: Is {\tt CCC1=[O+][Cu-3]2([O+]=C(CC)C1)[O+]=C(CC) CC(CC)=[O+]2} an HIV inhibitor?

\textit{output text}: 0

\noindent
\textbf{Example on BACE classification task:} 

\textit{input text with prompt}: Please evaluate the ability of {\tt S(=O)(=O)(CCCCC)C[C@@H](NC(=O)c1cccnc1)C(=O) N[C@H]([C@H](O)C[NH2+]Cc1cc(ccc1)CC)Cc1cc(F)cc(F)c1} to inhibit human beta-secretase

\textit{output text}: 1

\noindent
\textbf{Example on logS prediction task:} 

\textit{input text with prompt}: Given molecule with SMILES {\tt OCC2OC(Oc1ccccc1CO)C(O)C(O)C2O}, predict its logS

\textit{output text}: 1.083897

\noindent
\textbf{Example on HFE prediction task:} 

\textit{input text with prompt}: What hydration free energy does {\tt COc1cc(c(c(c1O)OC)Cl)C=O} have?

\textit{output text}: -1.013714

\noindent
\textbf{Example on logD prediction task:} 

\textit{input text with prompt}: What is the lowest unoccupied molecular orbital (LUMO) energy of this molecule? : {\tt O=C1OC2C3CC1OC32 }

\textit{output text}: 0.0035

\noindent
\textbf{Example on HOMO-LUMO prediction task:} 

\textit{input text with prompt}: lipophilic is {\tt COc1cc(OC)c(cc1NC(=O)CCC(=O)O)S(=O)\\(=O)NCc2ccccc2N3CCCCC3}?

\textit{output text}: -0.720000

\subsection{NLP: Tasks and Datasets}\label{sec:datasets}

\subsubsection{Named entity recognition}

Named entity recognition (NER) is a fundamental aspect of natural language processing, involving the identification and classification of entities in a given text into predefined categories. In biomedical NER, the focus lies in extracting mentions of diseases, genes, chemicals, and other biologically relevant entity types. To conduct this study, we carefully selected five datasets: 
\begin{itemize}
    \item BC2GM \cite{smith2008biocreative};
    \item BC5CDR-Disease \cite{li2016biocreative};
    \item BC5CDR-Chemical \cite{li2016biocreative};
    \item JNLPBA \cite{collier2004introduction};
    \item NCBI-Disease \cite{dougan2014ncbi}.
\end{itemize}

\paragraph{BC2GM}
The BC2GM dataset encompasses an extensive collection of over 20,000 sentences extracted from the MEDLINE database, spanning the years 1991 to 2003. Each document in this dataset is annotated with gene mention spans, amounting to a total of 24,583 mentions.

\paragraph{BC5CDR}
The BioCreative V CDR dataset was specifically designed for named entity recognition tasks involving disease and chemical entity types. It contains 12,850 disease and 15,935 chemical mentions, drawn from 1,500 PubMed articles.

\paragraph{JNLPBA}
The JNLPBA involves gene mention annotations across more than 2,000 PubMed abstracts. The creation of this dataset entailed a meticulous search on the MEDLINE database, using specific MeSH terms such as 'human', 'blood cells', and 'transcription factors'. In total, JNLPBA comprises 59,963 gene mention spans.

\paragraph{NCBI-Disease}
The NCBI-disease corpus, developed by the National Center for Biotechnology Information (NCBI), constitutes a vast collection of 793 PubMed abstracts that have undergone meticulous annotation by domain experts. These annotations include disease names and their corresponding concept IDs, sourced from the Medical Subject Headings (MeSH) vocabulary \cite{lipscomb2000medical}.

%\paragraph{Technical Details}
In order to train the neural network in a text-to-text format, we designed five prompts. Each prompt asks to highlight the spans corresponding to mentions of specific entity. In order to achieve this, we insert specific tokens before and after the mention of an entity in the text. 

\noindent
\textbf{Evaluation metric:} the evaluation of the NER task's quality is performed using the entity level F-measure.

\noindent
\textbf{Example:} 

\textit{input text with prompt}: Please find all instances of diseases in the given text. Each mention should be surrounded by "diso*" and "*diso": Identification of APC2, a homologue of the adenomatous polyposis coli tumor suppressor; 

\textit{output text}: Identification of APC2 , a homologue of the diso* adenomatous polyposis coli tumour *diso suppressor.

\subsubsection{Question Answering}
Question Answering (QA) is an important area of NLP research. The objective of QA is to develop intelligent systems that can understand and accurately answer questions posed in natural language. Within the biomedical domain, QA refers to the specific applications and models designed to address questions related to biomedical and healthcare information. It is required for model to understand and respond to questions pertaining to medical knowledge, clinical data, scientific literature, drug information, and other relevant biomedical topics. In this study, we conducted experiments on four biomedical QA datasets:

\begin{itemize}
    \item BioASQ \cite{nentidis2020results};
    \item PubMedQA \cite{jin2019pubmedqa};
    \item MedMCQA \cite{pal2022medmcqa};
    \item MMLU \cite{hendrycks2020measuring}.
\end{itemize}
The first two datasets are employed to evaluate the neural network's ability to answer binary Yes/No questions, while the remaining two datasets are used in scenarios that involve multi-choice and open question answering.

\paragraph{BioASQ and PubMedQA}
BioASQ (Biomedical Question Answering) is a widely recognized dataset in the biomedical domain, specifically designed for evaluating question answering systems. Following the \cite{gu2021domain} we restrict the dataset to yes/no questions. We use the official train/dev/test split where each contains 670/75/140 questions respectively.

Similar to BioASQ, the PubMedQA dataset as well presents questions with limited number of answers. In contrast to the previous dataset, the answers to the questions in PubMedQA are selected from yes, no, or maybe. We use the original train/dev/test split with 450, 50, and 500 questions, respectively.

\paragraph{MedMCQA and MMLU}
For multiple choice question answering, we employ the concatenation of the MedMCQA and MMLU datasets from \cite{fang2023mol}, resulting in a total of 12,398 multiple-choice questions. As \cite{fang2023mol} does not provide train/dev/test partitions, we randomly split the dataset into a ratio of 75:25.

To perform open question answering, we adopted a dataset introduced in \cite{fang2023mol}, which comprises 27,574 question-answer pairs. This dataset was curated from the MedMCQA dataset.

%\paragraph{Technical Details} 

\textbf{Evaluation metric:} to evaluate the performance of yes/no and multiple-choice question-answering tasks, we utilized the accuracy metric. For open-ended question-answering tasks, we adopted the BLEU-2 metric as our evaluation criterion.

\noindent
\textbf{Yes/No QA example:} 

\textit{input text with prompt}: Given a passage: De novo DNA methylation in Arabidopsis thaliana is catalyzed by the methyltransferase DRM2, a homolog of the mammalian de novo methyltransferase DNMT3. Here we describe DNA methyltransferase genes from both Arabidopsis and maize that show a high level of sequence similarity to Dnmt3, suggesting that they encode plant de novo methyltransferases. Relative to all known eukaryotic methyltransferases, these plant proteins contain a novel arrangement of the motifs required for DNA methyltransferase catalytic activity. The N termini of these methyltransferases contain a series of ubiquitin-associated (UBA) domains.  BLASTX searches and phylogenetic analysis suggested that five cDNAs belonged to four classes (Dnmt1, Dnmt2, CMT and Dnmt3) of DNA methyltransferase genes, 

answer the question: Are there any DNMT3 proteins present in plants?;  

\textit{output text}: Yes.

\noindent
\textbf{Multi-choice QA example:} 

\textit{input text with prompt}: Which of the following is antifibrinolytic drug: What of the following is the right choice?

(A) Tenecteplase

(B) Heparin

(C) Urokinase 

(D) Tranexaemic acid

\textit{output text}: The final answer is (D).

\noindent
\textbf{Open ended QA example:}

\textit{input text with prompt}: 1,25 dihydrocholecalciferol acts on?

\textit{output text}: Intranuclear receptors

\subsubsection{Relation Extraction}

Relation extraction (RE) is a NLP task that involves identifying and classifying the relationships between entities mentioned in a text. In the biomedical domain, RE refers to the specific application of RE techniques and models to extract and classify relationships between biomedical entities mentioned in text. Biomedical RE focuses on identifying and categorizing the associations between various biomedical entities, including genes, proteins, diseases, drugs, and other molecular entities. For experiments, we use three corpora:
\begin{itemize}
    \item ChemProt \cite{krallinger2017overview};
    \item DDI \cite{herrero2013ddi};
    \item GAD \cite{bravo2015extraction}.
\end{itemize}

\paragraph{ChemProt}
The ChemProt dataset is a widely used benchmark for the task of chemical-protein RE. The dataset comprises PubMed abstracts that are annotated with chemical-protein interactions, where the chemicals typically represent drug compounds or small molecules, and the proteins denote specific biological targets or enzymes. Each annotated interaction is labeled with the corresponding chemical and protein mentions, along with the following types of relationship: upregulator, downregulator, antagonist, agonist, and substrate. The training set of the dataset contains 9,995 relation pairs, and the test set contains 5,744 relation pairs.

\paragraph{DDI}
The DDI (Drug-Drug Interaction) corpus is a dataset designed for the purpose of identifying drug-drug interactions mentioned in biomedical texts. The corpus consists of annotated sentences or text passages that describe interactions between pairs of drugs. Each annotated interaction is labeled with the names of the drugs involved and the specific type of interaction. We employ the train/test split produced in \cite{gu2021domain}, where the training set contains 4,021 relation pairs and the test set contains 979 relation pairs.

\paragraph{GAD}
The GAD dataset is a comprehensive collection of genetic association information that was semi-automatically compiled using the Genetic Association Archive. In our study, we utilize an existing preprocessed version of GAD and its corresponding train/test split, which was created by Lee et al. \cite{lee2020biobert}. The training set of the dataset consists of 4,796 relation pairs, while the testing set includes 534 relation pairs.

%\paragraph{Technical details}
In our experimental framework, we adopt a binary classification approach for relation extraction. Here, the positive class indicates the presence of the specified type of relationship between two entities. 

\noindent
\textbf{Evaluation metric:} to evaluate the quality of RE tasks we utilize the F-1 measure of positive class.

\noindent
\textbf{Example:} 

\textit{input text with prompt}: does the Chlorprothixene and lithium are said to have mechanism type of interaction in the following passage:

Chlorprothixene may increase the plasma-level of concomitantly given lithium. In order to avoid lithium intoxication, lithium plasma levels should be monitored closely. If chlorprothixene is given concomitantly with opioids, the opioid dose should be reduced (by approx. 50\%), because chlorprothixene amplifies the therapeutic actions and side-effects of opioids massively. Avoid the concomitant use of chlorprothixene and tramadol (Ultram). Massive seizures may be encountered with this combination. Consider additive sedative effects and confusional states to emerge, if chlorprothixene is given with benzodiazepines or barbituates. Choose particular low doses of these drugs. Exert particular caution in combining chlorprothixene with other anticholinergic drugs (tricyclic antidepressants and antiparkinsonian agents): Particularly the elderly may develop delirium, high fever, severe obstipation, even ileus and glaucoma.

\textit{output text}: Yes

\subsubsection{Textual Entailment}

Textual entailment (TE) is a natural language processing task that involves determining the logical relationship between two pieces of text: a text fragment known as the "premise" and another text fragment known as the "hypothesis." The task is to decide whether the meaning of the hypothesis can be logically inferred or entailed from the meaning of the premise. For conducting our experiments, we utilize the following corpora:
\begin{itemize}
    \item MedNLI \cite{shivade2019mednli};
    \item SciTail \cite{khot2018scitail};
\end{itemize}

\paragraph{MedNLI}
MedNLI (Medical Natural Language Inference) is a specialized dataset designed to facilitate research in natural language inference within the medical and healthcare domain. It consists of pairs of sentences, where each pair comprises a premise and a hypothesis. The premise represents a clinical or biomedical context, while the hypothesis is a medical statement or claim that may or may not logically follow from the premise. Each sentence pair is annotated with one of three labels: "entailment," indicating that the hypothesis can be logically inferred from the premise; "contradiction," suggesting that the hypothesis contradicts the information in the premise; and "neutral," signifying that there is no logical relationship between the two sentences. The dataset comprises a total of 12,627 sentence pairs in the training set and 1,422 sentence pairs in the testing set.

\paragraph{SciTail}
The SciTail dataset is similar to the MedNLI dataset was designed for the task of natural language inference. Except that it covers a broader scientific domain. The train part of the corpora contains 24900 sentence pairs and the test part of the corpora contains 2126. 

%\paragraph{Technical Details} 
%\noindent
\textbf{Evaluation metric:} to evaluate the quality of TE tasks we utilize the Accuracy score.

\noindent
\textbf{Example:} 

\textit{input text with prompt}:
Given that "At [**Hospital 1456**] Hospital the patient was experiencing 10 out of 10 chest pain and received nitropaste two inches, three sublingual nitroglycerins, morphine 4 mg intravenously, Lopressor 5 mg intravenously." Does it follow that " The patient is asymptomatic."

yes or no?

\textit{output text}: No

\subsubsection{Sentence similarity}

Textual similarity tasks in the biomedical domain involve assessing the degree of semantic similarity or relatedness between pairs of biomedical texts. The goal of these tasks is to determine how closely two pieces of text, such as sentences or documents, are semantically or conceptually aligned. To conduct our experiments, we employ the BIOSSES dataset \cite{souganciouglu2017biosses}. 

\paragraph{BIOSSES}
The BIOSSES (Biomedical Sentence Similarity Benchmark) dataset is a specialized dataset designed to evaluate sentence similarity models in the biomedical domain. It contains pairs of biomedical sentences that are carefully selected to represent different levels of semantic similarity. Each sentence pair is annotated with a similarity score that represents the degree of semantic relatedness between the two sentences. The scores are typically on a continuous scale, indicating how similar or dissimilar the sentences are in meaning. The dataset comprises a total of 80 sentence pairs in the training set and 20 sentence pairs in the testing set.

%\paragraph{Technical Details} 
%\noindent
\textbf{Evaluation metric:} to evaluate the quality of Textual Similarity tasks we utilize the Pearson corellation score.

\noindent
\textbf{Example:} 

\textit{input text with prompt}:
Please assess the similarity between these two sentences on a scale of 0.0 (lowest) to 4.0 (highest). First sentence: "It has recently been shown that Craf is essential for Kras G12D-induced NSCLC." Second sentence:"It has recently become evident that Craf is essential for the onset of Kras-driven non-small cell lung cancer. "

\textit{output text}: 4.0

\subsubsection{Document Classification}
In the biomedical domain, the document classification task involves categorizing entire documents, such as scientific articles, research papers, or clinical reports, into predefined categories or classes. The goal is to automatically assign each document to the most relevant category based on its content and subject matter. For our experimental purposes, we utilize the Hallmarks of Cancer dataset.

\paragraph{Hallmarks of Cancer}

The Hallmarks of Cancer (HoC) dataset serves as a document classification task, centered around the concept of cancer hallmarks as established in the referenced work \cite{hanahan2000hallmarks}. This corpus comprises PubMed abstracts, each labeled with binary annotations, denoting the presence of specific discussions related to individual cancer hallmarks. We utilize the train/test split from \cite{gu2021domain} which comprises 13917 sentences in train part and 3547 sentences in test part. 

%\paragraph{Technical Details} 
%\noindent
\textbf{Evaluation metric:} to evaluate the quality of Document Classification tasks we utilize the F-1 score.

\noindent
\textbf{Example:} 

\textit{input text with prompt}:
Pick one category for the following text. The options are - activating invasion and metastasis, avoiding immune destruction, cellular energetics, enabling replicative immortality, evading growth suppressors, genomic instability and mutation, inducing angiogenesis, resisting cell death, none, sustaining proliferative signaling, tumor promoting inflammation.

Biopsy of a skin lesion showed lymphoproliferative infiltration of the dermis with a follicular and angiocentric growth pattern and regional epidermal necrosis.

\textit{output text}: resisting cell death

\subsubsection{PICO extraction}
PICO extraction is an essential NLP task that aims to automatically identify and extract specific fragments of text pertaining to the Patient (P), Intervention (I), Comparator (C), and Outcome (O) elements from unstructured biomedical texts, such as research articles and clinical trial reports. Typically, Comparator labels are omitted from the annotations, as they conform to established clinical trial norms, with "placebo" as the passive control and "standard of care" as the active control. To conduct our study, we leveraged the EBM PICO \cite{nye2018corpus} dataset for this purpose.

\paragraph{EBM PICO}

The EBM PICO dataset was specifically created to facilitate PICO extraction tasks. It employs token-level labeling, where each token is categorized into one of the PIO classes (Patient, Intervention, Outcome). The dataset comprises a total of 4,800 labeled abstracts for training purposes and 200 labeled abstracts for testing purposes.

%\paragraph{Technical Details} 
To conduct the PICO extraction task in a text-to-text format, we adopted the same prompt style as used for the Named Entity Recognition (NER) dataset.

\noindent
\textbf{Evaluation metric:} to evaluate the quality of PICO extraction tasks we utilize the word-level F-1 score.

\noindent
\textbf{Example:} 

\textit{input text with prompt}:
Please find all instances of Interventions in the given text. Each mention should be surrounded by "Intervention*" and "*Intervention": Study protocol : Rehabilitation including Social and Physical activity and Education in Children and Teenagers with Cancer ( RESPECT ) 

\textit{output text}: Study protocol :  Intervention* Rehabilitation including Social and Physical activity and Education *Intervention  in Children and Teenagers with Cancer ( RESPECT ) .

\balance

%If notes are included in your references you can change the title from 'References' to 'Notes and references' using the following command:
%\renewcommand\refname{Notes and references}

%%%REFERENCES%%%
\bibliography{rsc} %You need to replace "rsc" on this line with the name of your .bib file
\bibliographystyle{rsc} %the RSC's .bst file

\end{document}